\documentclass{article}
\usepackage{amssymb}
\usepackage{makeidx}
\usepackage{amsfonts}
\usepackage{amsmath}
\usepackage{grey}
\usepackage{color}
\usepackage{framed}
\usepackage{float}
\usepackage{graphicx}
\usepackage{amscd}
\usepackage{hyperref}
\usepackage[margin=3cm]{geometry}%
\input{tcilatex}
\begin{document}

\thispagestyle{empty}

\begin{center}
\begin{tabular}{c}
{\Large Minimum Relative Entropy Inference} \\ 
{\Large for Normal and Monte Carlo Distributions} \\ 
\\ 
\\ 
{\large Marcello Colasante} \\ 
marcello.colasante@arpm.co \\ 
\\ 
{\large Attilio Meucci} \\ 
attilio.meucci@arpm.co \\ 
\\ 
\\ 
this revision:\ February 7, 2020%
\end{tabular}

\bigskip\bigskip

\bigskip

\bigskip

\bigskip

\bigskip

\bigskip

\bigskip

\textbf{Abstract}
\end{center}

\noindent We represent affine sub-manifolds of exponential family
distributions as minimum relative entropy sub-manifolds. With such
representation we derive analytical formulas for the inference from partial
information on expectations and covariances of multivariate normal
distributions; and we improve the numerical implementation via Monte Carlo
simulations for the inference from partial information of generalized
expectation type.\bigskip

\bigskip

\bigskip

\newpage

\section{Introduction\label{intro}}

Inference is ubiquitous in financial applications: stress-testing and
scenario analysis, such as in \cite{MinaXiao01}, explore the consequences of
specific market scenarios on the distribution of the portfolio loss.\
Similar, portfolio construction techniques such as \cite{BlackLitt90} inject
views on specific factor returns into the estimated distribution of a broad
market.

A general approach to perform inference under partial information based on
the principle of minimum relative entropy (MRE)\ was explored in \cite%
{Meucci08a}. In the original paper, the general theory was supported by two
applications:\ an analytical solution under normality, and a numerical
algorithm for distributions represented by scenarios, such as Monte Carlo,
historical, or categorical.

Here we enhance both the analytical and the numerical implementations of 
\cite{Meucci08a} drawing from results in \cite{Colas19}.

In Section \ref{preliminaries} we state well-known results to set the
notation and background.

In Section \ref{sec-normal-assumption}, we embed the analytical MRE problem
under normality and information on expectations and covariances of arbitrary
linear combinations into a broader analytical framework. In computing the
solution, we find that the updated expectation in \cite{Meucci08a} must be
adjusted by a term implied by the information on the covariances.

In Section \ref{scen-prob-sec}, we address the MRE problem numerically. Most
numerical applications of MRE which involve Monte Carlo sampling methods,
such as stochastic approximation, or sample path optimization algorithms,
see \cite{Schofield07}, could be inefficient. On the other hand, the
scenario-based MRE\ algorithm in \cite{Meucci08a} does not entail drawing
scenarios, and as such is efficient, but subject to the curse of
dimensionality which may affect precision. Here we improve the original
scenario-based MRE\ in \cite{Meucci08a} with an iterative procedure based on
Hamiltonian Monte Carlo sampling \cite{Chao15}, \cite{Neal11}, thereby
achieving more precision.

In Section \ref{sec-application} we present a case study that applies and
compares the analytical solution and the numerical algorithm.

Finally, in Section \ref{conclusions-1} we list the main contributions.

\section{Background\label{preliminaries}}

In this section we briefly review well-known results, refer to \cite%
{Jaakkola99}, \newline
\cite{CoverThomas06}, \cite{AmariNagaoka00}, \cite{AmariShun16} for more
details.

Let $\boldsymbol{X}\equiv\left( X_{1},\ldots,X_{\bar{n}}\right) ^{\prime}$
be a target vector with a reference base distribution with support $\mathcal{%
X}\subseteq\mathbb{R}^{\bar{n}}$, as represented by the probability density
function (pdf)%
\begin{equation}
\boldsymbol{X}\sim\underline{f}_{\boldsymbol{X}}\text{,}  \label{prior-distr}
\end{equation}
that needs to be estimated via historical, maximum likelihood, GMM etc. Let $%
\boldsymbol{Z}\equiv\left( Z_{1},\ldots,Z_{\bar{k}}\right) ^{\prime}$ be a
random vector of inference input variables, on which we have new
information. Without loss of generality, we can assume that the inference
input variables are transformation of the target variables%
\begin{equation}
\boldsymbol{Z}\equiv\zeta(\boldsymbol{X})\text{,}  \label{view-varr}
\end{equation}
for a suitable multivariate function $\zeta:\mathbb{R}^{\bar{n}}\rightarrow%
\mathbb{R}^{\bar{k}}$. In applications, the number $\bar{n}$ of target
variables is typically much larger than the number $\bar{k}$ of inference
variables%
\begin{equation}
\bar{k}\ll\bar{n}\text{.}  \label{less-view-mkt}
\end{equation}
Inference amounts to assessing the impact of some information, or subjective
views, on the distribution of $\boldsymbol{X}$, which can be expressed as
constraints on the distribution of the inference variables 
\begin{equation}
f_{\boldsymbol{Z}}\in\mathcal{C}_{\boldsymbol{Z}}\text{,}
\label{part-view-pos}
\end{equation}
which in general are violated by the base distribution (\ref{prior-distr}).

The principle of minimum relative entropy (MRE) is a standard approach to
inference with partial information.\ Let us denote the relative entropy
between distributions as follows%
\begin{equation}
\mathcal{E(}f_{\boldsymbol{X}}\Vert\underline{f}_{\boldsymbol{X}})\equiv
\int_{\mathcal{X}}f_{\boldsymbol{X}}\left( \boldsymbol{x}\right) \ln (\frac{%
f_{\boldsymbol{X}}\left( \boldsymbol{x}\right) }{\underline{f}_{\boldsymbol{X%
}}\left( \boldsymbol{x}\right) })d\boldsymbol{x}\text{.}  \label{rel-entropy}
\end{equation}
Then, according to the MRE, the updated inferred distribution is the closest
to the base $\underline{f}_{\boldsymbol{X}}$ (\ref{prior-distr})%
\begin{equation}
\bar{f}_{\boldsymbol{X}}\equiv\limfunc{argmin}\limits_{f_{\boldsymbol{X}}\in%
\mathcal{C}_{\boldsymbol{X}}}\mathcal{E}(f_{\boldsymbol{X}}\Vert \underline{f%
}_{\boldsymbol{X}})\text{,}  \label{mre-posterior}
\end{equation}
which at the same time satisfies the information constraints (\ref%
{part-view-pos}) induced by the inference variables, or $\mathcal{C}_{%
\boldsymbol{X}}\equiv\{f_{\boldsymbol{X}}:\quad f_{\boldsymbol{Z}}\in%
\mathcal{C}_{\boldsymbol{Z}}\}$.

In particular, here we consider information (\ref{part-view-pos}) expressed
in terms of expectation 
\begin{equation}
\mathcal{C}_{\boldsymbol{X}}\equiv\{f_{\boldsymbol{X}}:\quad\mathbb{E}^{f_{%
\boldsymbol{X}}}\{\zeta(\boldsymbol{X})\}=\boldsymbol{\eta}^{\mathit{info}}\}%
\text{,}  \label{view-gen-exp-norm}
\end{equation}
where $\boldsymbol{\eta}^{\mathit{info}}\equiv(\eta_{1}^{\mathit{info}%
},\ldots,\eta_{\bar{k}}^{\mathit{info}})^{\prime}$ is a $\bar{k}\times1$
vector and $\zeta$ is an arbitrary function. The equality conditions (\ref%
{view-gen-exp-norm}) cover a wide range of practical applications, such as
information on volatilities, correlations, tail behaviors, etc. More general
inequality constraints $\mathbb{E}^{f_{\boldsymbol{X}}}\{\zeta (\boldsymbol{X%
})\}\leqq\boldsymbol{\eta}^{\mathit{info}}$ are also tractable, but beyond
the scope of this article.

Then the MRE updated distribution (\ref{mre-posterior}) belongs to the
exponential family class%
\begin{equation}
\bar{f}_{\boldsymbol{X}}\quad\Leftrightarrow\quad\mathit{Exp}(\boldsymbol{%
\theta}^{\mathit{info}},\zeta,\underline{f}_{\boldsymbol{X}},\mathcal{X})%
\text{,}  \label{exp-family-set}
\end{equation}
which means the pdf reads 
\begin{equation}
\bar{f}_{\boldsymbol{X}}=\underline{f}_{\boldsymbol{X}}(\boldsymbol{x})e^{%
\boldsymbol{\theta}^{\mathit{info}\prime}\zeta(\boldsymbol{x})-\psi(%
\boldsymbol{\theta}^{\mathit{info}})}\text{,}  \label{exp-family-set-1}
\end{equation}
where $\psi(\boldsymbol{\theta})$ is the log-partition function%
\begin{equation}
\psi(\boldsymbol{\theta})\equiv\ln\int_{\mathcal{X}}e^{\boldsymbol{\theta }%
^{\prime}\zeta(\boldsymbol{x})}\underline{f}_{\boldsymbol{X}}\left( 
\boldsymbol{x}\right) d\boldsymbol{x}\text{.}  \label{soidfodngf}
\end{equation}
According to (\ref{exp-family-set}) the sufficient statistics $\zeta (%
\boldsymbol{x})$ are the information functions specifying the inference
input variables (\ref{view-varr}); the expectation parameters $\boldsymbol{%
\eta }^{\mathit{info}}$ are the features quantifying the information
constraints (\ref{view-gen-exp-norm}); and the natural parameters $%
\boldsymbol{\theta }^{\mathit{info}}\equiv(\theta_{1}^{\mathit{info}%
},\ldots,\theta_{\bar{k}}^{\mathit{info}})^{\prime}$ are the Lagrange
multipliers of the MRE problem (\ref{mre-posterior})-(\ref{view-gen-exp-norm}%
), which are related to the expectation parameters $\boldsymbol{\eta}^{%
\mathit{info}}$ via the Legendre transform of the log-partition, or link
function%
\begin{equation}
\boldsymbol{\theta}^{\mathit{info}}\equiv\nabla\psi^{-1}(\boldsymbol{\eta }^{%
\mathit{info}})\text{.}  \label{lagr-prob}
\end{equation}

The key to obtain the MRE\ updated distribution (\ref{exp-family-set}) are
the Lagrange multipliers (\ref{lagr-prob}). However solving (\ref{lagr-prob}%
) is not feasible in general.

\section{Analytical results\label{sec-normal-assumption}}

To obtain analytical results, we make two further assumptions:

\begin{itemize}
\item The base distribution (\ref{prior-distr}) is of an exponential family
class%
\begin{equation}
\underline{f}_{\boldsymbol{X}}\quad\Leftrightarrow\quad\mathit{Exp}(%
\underline{\boldsymbol{\theta}}_{\boldsymbol{X}},\tau,h,\mathcal{X})\text{,}
\label{exp-fam-base}
\end{equation}
for a reference measure $h(\boldsymbol{x})$, natural parameters $\underline{%
\boldsymbol{\theta}}_{\boldsymbol{X}}\equiv(\underline{\theta }_{\boldsymbol{%
X};1},\ldots,\underline{\theta}_{\boldsymbol{X};\bar{l}})^{\prime}$ within a
parameter domain $\Theta\subseteq\mathbb{R}^{\bar{l}}$, sufficient
statistics $\tau(\boldsymbol{x})\equiv(\tau_{1}(\boldsymbol{x}),\ldots,\tau_{%
\bar{l}}(\boldsymbol{x}))^{\prime}$.

\item The information is of expectation type (\ref{view-gen-exp-norm}) and
linear in the sufficient statistics%
\begin{equation}
f_{\boldsymbol{X}}:\quad\mathbb{E}^{f_{\boldsymbol{X}}}\{\boldsymbol{\gamma }%
\tau(\boldsymbol{X})\}=\boldsymbol{\eta}^{\mathit{info}}\text{,}
\label{quad-view-func-norm}
\end{equation}
for a $\bar{k}\times\bar{l}$ matrix $\boldsymbol{\gamma}$.
\end{itemize}

Then, the MRE updated distribution (\ref{exp-family-set}) is a
\textquotedblleft curved\textquotedblright\ sub-family of the same
exponential family class as the base [\ref{gen-post-pdf-appendix}]%
\begin{equation}
\bar{f}_{\boldsymbol{X}}\quad\Leftrightarrow\quad\mathit{Exp}(\boldsymbol{%
\bar {\theta}}_{\boldsymbol{X}},\tau,h,\mathcal{X})\text{,}
\label{exp-fam-conj-update}
\end{equation}
where the new natural parameters are an affine transformation (and thus not
literally \textquotedblleft curved\textquotedblright) of the\ optimal
Lagrange multipliers $\boldsymbol{\theta}^{\mathit{info}}$%
\begin{equation}
\boldsymbol{\bar{\theta}}_{\boldsymbol{X}}\equiv\underline{\boldsymbol{%
\theta }}_{\boldsymbol{X}}+\boldsymbol{\gamma}^{\prime}\boldsymbol{\theta }^{%
\mathit{info}}\text{,}  \label{post-can-param-norm}
\end{equation}
as long as $\boldsymbol{\bar{\theta}}_{\boldsymbol{X}}\in\Theta$.

\subsection{Categorical distribution\label{sec-normal-assumption-cat}}

For a trivial example of the result (\ref{post-can-param-norm}), let us
consider for the base (\ref{prior-distr}) a scenario-probability
distribution (or generalized categorical distribution) $\boldsymbol{X}\sim\{%
\boldsymbol{x}^{(j)},\underline{p}^{\left( j\right) }\}_{j=1}^{\bar{\jmath}}$%
, which belongs to a specific exponential family class (\ref{exp-fam-base}) 
\begin{equation}
\{\boldsymbol{x}^{(j)},\underline{p}^{\left( j\right) }\}_{j=1}^{\bar {\jmath%
}}\quad\Leftrightarrow\quad\mathit{Exp}(\{\ln\frac{\underline{p}^{(j)}}{%
\underline{p}^{(\bar{\jmath})}}\}_{j=1}^{\bar{\jmath}-1},\{1_{\boldsymbol{x}=%
\boldsymbol{x}^{\left( j\right) }}\}_{j=1}^{\bar{\jmath}-1},1,\{\boldsymbol{x%
}^{(j)}\}_{j=1}^{\bar{\jmath}})\text{,}  \label{scen-prob-deff}
\end{equation}
where $\boldsymbol{x}^{\left( j\right) }$ are $\bar{\jmath}$ joint scenarios
for $\boldsymbol{X}$; the canonical parameters are the multi-logit
transformation of the scenarios probabilities $\underline{p}^{\left(
j\right) }\equiv\underline{\mathbb{P}}\{\boldsymbol{X}=\boldsymbol{x}%
^{\left( j\right) }\}$, which are positive and sum to one; and the
sufficient statistics are the one-hot encoding functions, see e.g. \cite%
{AmariShun16}. In this framework, \emph{any} expectation conditions as in (%
\ref{view-gen-exp-norm}) can be expressed as linear statements in the
sufficient statistics (\ref{post-can-param-norm})%
\begin{equation}
f_{\boldsymbol{X}}:\quad\mathbb{E}^{f_{\boldsymbol{X}}}\{\tsum
\nolimits_{j=1}^{\bar{\jmath}-1}\boldsymbol{\gamma}^{\left( j\right) }1_{%
\boldsymbol{X}=\boldsymbol{x}^{\left( j\right) }}\}=\boldsymbol{\eta}^{%
\mathit{info}}-\zeta (\boldsymbol{x}^{(\bar{\jmath})})\text{,}
\label{one-hot-sufficient-1}
\end{equation}
where $\boldsymbol{\gamma}^{\left( j\right) }\equiv\zeta(\boldsymbol{x}%
^{(j)})-\zeta(\boldsymbol{x}^{(\bar{\jmath})})$.

Then, from (\ref{exp-fam-conj-update}), the MRE updated distribution (\ref%
{exp-family-set}) must be a scenario-probability distribution as the base (%
\ref{scen-prob-deff})%
\begin{equation}
\{\boldsymbol{x}^{(\bar{\jmath})},\bar{p}^{(j)}\}_{j=1}^{\bar{\jmath}%
}\quad\Leftrightarrow\quad\mathit{Exp}(\{\ln\frac{\bar{p}^{(j)}}{\bar{p}^{(%
\bar{\jmath})}}\}_{j=1}^{\bar{\jmath}-1},\{1_{\boldsymbol{x}=\boldsymbol{x}%
^{\left( j\right) }}\}_{j=1}^{\bar{\jmath}-1},1,\{\boldsymbol{x}%
^{(j)}\}_{j=1}^{\bar{\jmath}})\text{,}  \label{one-hot-sufficient-2}
\end{equation}
but with new probabilities $\bar{p}^{(j)}$, as follows from (\ref%
{post-can-param-norm})%
\begin{equation}
\ln\frac{\bar{p}^{(j)}}{\bar{p}^{(\bar{\jmath})}}=\ln\frac{\underline{p}%
^{(j)}}{\underline{p}^{(\bar{\jmath})}}+\boldsymbol{\gamma}^{\left( j\right)
\prime}\boldsymbol{\theta}^{\mathit{info}}\text{,}
\label{one-hot-sufficient-3}
\end{equation}
for any $j=1,\ldots,\bar{\jmath}-1$. This leads to the numerical MRE
algorithm for scenario-probability distributions in \cite{Meucci08c}, which
we use in Section \ref{scen-prob-sec}.

\subsection{Normal distribution\label{sec-normal-assumption-norm}}

For a non-trivial instance of the result (\ref{post-can-param-norm}), let us
consider the special case of (\ref{exp-fam-base})-(\ref{quad-view-func-norm}%
)\ that generalizes the parametric MRE in \cite{Meucci08c} and corrects an
error therein.

More precisely, let us assume that the base (\ref{prior-distr}) is a normal
distribution, which belongs to a specific exponential family class (\ref%
{exp-fam-base}) 
\begin{equation}
\mathit{N}(\underline{\boldsymbol{\mu}}_{\boldsymbol{X}},\underline{%
\boldsymbol{\sigma}}_{\boldsymbol{X}}^{2})\quad\Leftrightarrow \quad\mathit{%
Exp}(\underline{\boldsymbol{\theta}}_{\boldsymbol{X}}^{\mathit{N}},\tau^{%
\mathit{N}},(2\pi)^{-\bar{n}/2},\mathbb{R}^{\bar{n}})\text{,}
\label{norm-exp-form}
\end{equation}
where the canonical coordinates are suitable transformations of the $\bar {n}%
\times1$ expectation vector $\underline{\boldsymbol{\mu}}_{\boldsymbol{X}}$
and the $\bar{n}\times\bar{n}$ covariance matrix $\underline{\boldsymbol{%
\sigma}}_{\boldsymbol{X}}^{2}$%
\begin{equation}
\underline{\boldsymbol{\theta}}_{\boldsymbol{X}}^{\mathit{N}}\equiv\left( 
\begin{array}{c}
\underline{\boldsymbol{\theta}}_{\boldsymbol{X};\mu}^{\mathit{N}} \\ 
\mathit{vec}(\underline{\boldsymbol{\theta}}_{\boldsymbol{X};\sigma,\sigma
}^{\mathit{N}})%
\end{array}
\right) \equiv\left( 
\begin{array}{c}
(\underline{\boldsymbol{\sigma}}_{\boldsymbol{X}}^{2})^{-1}\underline{%
\boldsymbol{\mu}}_{\boldsymbol{X}} \\ 
-\frac{1}{2}\mathit{vec}((\underline{\boldsymbol{\sigma}}_{\boldsymbol{X}%
}^{2})^{-1})%
\end{array}
\right) \text{;}  \label{norm-can-cord-pri}
\end{equation}
and where sufficient statistics are pure linear and quadratic functions%
\begin{equation}
\tau^{\mathit{N}}(\boldsymbol{x})\equiv\left( 
\begin{array}{c}
\tau_{\mu}^{\mathit{N}}(\boldsymbol{x}) \\ 
\tau_{\sigma,\sigma}^{\mathit{N}}(\boldsymbol{x})%
\end{array}
\right) \equiv\left( 
\begin{array}{c}
\boldsymbol{x} \\ 
\mathit{vec}(\boldsymbol{xx}^{\prime})%
\end{array}
\right) \text{.}  \label{norm-suff-stat-pri}
\end{equation}
Then let us consider MRE\ inference as in (\ref{mre-posterior})%
\begin{equation}
\bar{f}_{\boldsymbol{X}}\equiv\limfunc{argmin}\limits_{f_{\boldsymbol{X}}\in%
\mathcal{C}_{\boldsymbol{X}}}\mathcal{E}(f_{\boldsymbol{X}}\Vert \underline{f%
}_{\boldsymbol{X}})\text{,}  \label{mre-posterior-1}
\end{equation}
under information on linear combinations of expectations and covariances%
\begin{equation}
f_{\boldsymbol{X}}\in\mathcal{C}_{\boldsymbol{X}}:\qquad\left\{ 
\begin{array}{l}
\mathbb{E}^{f_{\boldsymbol{X}}}\{\boldsymbol{\gamma}_{\mu}\boldsymbol{X}\}=%
\boldsymbol{\mu}^{\mathit{info}} \\ 
\mathbb{C}v^{f_{\boldsymbol{X}}}\{\boldsymbol{\gamma}_{\sigma}\boldsymbol{X}%
\}=\boldsymbol{\sigma}^{2\mathit{info}}%
\end{array}
\right.  \label{view-exp-cov-norm}
\end{equation}
where $\boldsymbol{\gamma}_{\mu}$ is a $\bar{k}_{\mu}\times\bar{n}$
full-rank matrix; $\boldsymbol{\mu}^{\mathit{info}}$ is a $\bar{k}%
_{\mu}\times1$ vector; $\boldsymbol{\gamma}_{\sigma}$ is a $\bar{k}%
_{\sigma}\times\bar{n}$ full-rank matrix; and $\boldsymbol{\sigma}^{2\mathit{%
info}} $ is a $\bar{k}_{\sigma }\times\bar{k}_{\sigma}$ symmetric and
positive definite matrix.

The inference constraints in the MRE problem (\ref{mre-posterior-1}) are not
of expectation type (\ref{quad-view-func-norm}). However, we can use a
two-step approach to leverage this result.

First, we consider all the possible expectation constraints (\ref%
{quad-view-func-norm}) compatible with the information (\ref%
{view-exp-cov-norm}) 
\begin{equation}
f_{\boldsymbol{X}}\in\mathcal{C}_{\boldsymbol{X}}^{(\boldsymbol{\eta}%
_{\sigma })}:\qquad\left\{ 
\begin{array}{l}
\mathbb{E}^{f_{\boldsymbol{X}}}\{\boldsymbol{\gamma}_{\mu}\boldsymbol{X}\}=%
\boldsymbol{\mu}^{\mathit{info}} \\ 
\mathbb{E}^{f_{\boldsymbol{X}}}\{\boldsymbol{\gamma}_{\sigma}\boldsymbol{XX}%
^{\prime}\boldsymbol{\gamma}_{\sigma}\}=\boldsymbol{\sigma}^{2\mathit{info}}+%
\boldsymbol{\eta}_{\sigma}\boldsymbol{\eta}_{\sigma}^{\prime} \\ 
\mathbb{E}^{f_{\boldsymbol{X}}}\{\boldsymbol{\gamma}_{\sigma}\boldsymbol{X}%
\}=\boldsymbol{\eta}_{\sigma}%
\end{array}
\right.  \label{view-exp-cov-norm-1}
\end{equation}
for any $\bar{k}_{\sigma}\times1$ vector $\boldsymbol{\eta}_{\sigma}$; and
the related MRE\ optimization%
\begin{equation}
f_{\boldsymbol{X}}^{(\boldsymbol{\eta}_{\sigma})}\equiv\limfunc{argmin}%
\limits_{f_{\boldsymbol{X}}\in\mathcal{C}_{\boldsymbol{X}}^{(\boldsymbol{%
\eta }_{\sigma})}}\mathcal{E}(f_{\boldsymbol{X}}\Vert\underline{f}_{%
\boldsymbol{X}})\text{.}  \label{view-exp-cov-norm-2}
\end{equation}
Because of the expectation constraints (\ref{quad-view-func-norm}), for any $%
\boldsymbol{\eta}_{\sigma}$ the solution $f_{\boldsymbol{X}}^{(\boldsymbol{%
\eta}_{\sigma})}$must be normal due to (\ref{exp-fam-conj-update}), and we
can compute it analytically [\ref{norm-post-pdf-appendix}]%
\begin{equation}
f_{\boldsymbol{X}}^{(\boldsymbol{\eta}_{\sigma})}\quad\Leftrightarrow \quad%
\mathit{N}(\mu(\boldsymbol{\eta}_{\sigma}),\bar{\boldsymbol{\sigma}}_{%
\boldsymbol{X}}^{2})\text{,}  \label{view-exp-cov-norm-3}
\end{equation}
for a suitable function $\mu(\cdot)$ and same updated covariance matrix 
\begin{equation}
\boldsymbol{\bar{\sigma}}_{\boldsymbol{X}}^{2}\equiv \underline{\boldsymbol{%
\sigma}}_{\boldsymbol{X}}^{2}+\boldsymbol{\gamma }_{\sigma}^{\dag}(%
\boldsymbol{\sigma}^{2\mathit{info}}-\boldsymbol{\gamma }_{\sigma}\underline{%
\boldsymbol{\sigma}}_{\boldsymbol{X}}^{2}\boldsymbol{\gamma}%
_{\sigma}^{\prime})\boldsymbol{\gamma}_{\sigma}^{\dag \prime}\text{,}
\label{comp-quad-norm-sig-1}
\end{equation}
where $\boldsymbol{\gamma}_{\sigma}^{\dag}$ is a $\bar{k}_{\sigma}\times 
\bar{n}$ (right) pseudo-inverse matrix for $\boldsymbol{\gamma}_{\sigma}$%
\begin{equation}
\boldsymbol{\gamma}_{\sigma}^{\dag}\equiv\underline{\boldsymbol{\sigma}}_{%
\boldsymbol{X}}^{2}\boldsymbol{\gamma}_{\sigma}^{\prime}(\boldsymbol{\gamma }%
_{\sigma}\underline{\boldsymbol{\sigma}}_{\boldsymbol{X}}^{2}\boldsymbol{%
\gamma}_{\sigma}^{\prime})^{-1}\text{.}  \label{equiv-pseudo-inv}
\end{equation}

Second, we compute the optimal vector $\boldsymbol{\eta}_{\sigma }^{\mathit{%
info}}$ that minimizes the relative entropy%
\begin{equation}
\boldsymbol{\eta}_{\sigma}^{\mathit{info}}\equiv\limfunc{argmin}_{%
\boldsymbol{\eta}_{\sigma}}\mathcal{E}(f_{\boldsymbol{X}}^{(\boldsymbol{\eta 
}_{\sigma})}\Vert\underline{f}_{\boldsymbol{X}})\text{,}
\label{view-exp-cov-norm-4}
\end{equation}
which turns out to be a simple quadratic programming problem in $\boldsymbol{%
\eta}_{\sigma}$ [\ref{norm-post-pdf-quad-pol-2}]. Then the updated
distribution (\ref{mre-posterior-1}) must be normal as in (\ref%
{view-exp-cov-norm-3}) [\ref{norm-post-pdf-quad-pol-2}]%
\begin{equation}
\bar{f}_{\boldsymbol{X}}=f_{\boldsymbol{X}}^{(\boldsymbol{\eta}_{\sigma }^{%
\mathit{info}})}\quad\Leftrightarrow\quad\mathit{N}(\boldsymbol{\bar{\mu}}_{%
\boldsymbol{X}},\boldsymbol{\bar{\sigma}}_{\boldsymbol{X}}^{2})\text{,}
\label{pos-exp-nomr-1}
\end{equation}
with updated expectation as follows%
\begin{equation}
\boldsymbol{\bar{\mu}}_{\boldsymbol{X}}\equiv\mu(\boldsymbol{\eta}_{\sigma
}^{\mathit{info}})=\boldsymbol{\bar{\mu}}_{\boldsymbol{X};\sigma}+\bar{%
\boldsymbol{\gamma}}_{\mu}^{\dag}(\boldsymbol{\mu}^{\mathit{info}}-%
\boldsymbol{\gamma}_{\mu}\boldsymbol{\bar{\mu}}_{\boldsymbol{X};\sigma })%
\text{,}  \label{comp-quad-norm-mu-1}
\end{equation}
where $\bar{\boldsymbol{\gamma}}_{\mu}^{\dag}$ is a $\bar{k}_{\mu}\times 
\bar{n}$ (right) pseudo-inverse matrix for $\boldsymbol{\gamma}_{\mu}$%
\begin{equation}
\bar{\boldsymbol{\gamma}}_{\mu}^{\dag}\equiv\boldsymbol{\bar{\sigma}}_{%
\boldsymbol{X}}^{2}\boldsymbol{\gamma}_{\mu}^{\prime}(\boldsymbol{\gamma }%
_{\mu}\boldsymbol{\bar{\sigma}}_{\boldsymbol{X}}^{2}\boldsymbol{\gamma}_{\mu
}^{\prime})^{-1}\text{;}  \label{mu-pseudo-inv-pos}
\end{equation}
and where $\boldsymbol{\bar{\mu}}_{\boldsymbol{X};\sigma}$ is an $\bar {n}%
\times1$ vector defined as follows%
\begin{equation}
\boldsymbol{\bar{\mu}}_{\boldsymbol{X};\sigma}=\underline{\boldsymbol{\mu}}_{%
\boldsymbol{X}}+\boldsymbol{\gamma}_{\sigma}^{\dag}(\boldsymbol{\sigma }^{2%
\mathit{info}}(\boldsymbol{\gamma}_{\sigma}\underline{\boldsymbol{\sigma}}_{%
\boldsymbol{X}}^{2}\boldsymbol{\gamma}_{\sigma}^{\prime})^{-1}\boldsymbol{%
\gamma}_{\sigma}\underline{\boldsymbol{\mu}}_{\boldsymbol{X}}-\boldsymbol{%
\gamma}_{\sigma}\underline{\boldsymbol{\mu}}_{\boldsymbol{X}})\text{.}
\label{view-exp-cov-norm-4b}
\end{equation}

In the special case of uncorrelated information variables under the base
distribution (\ref{norm-exp-form})%
\begin{equation}
\mathbb{C}v^{\underline{f}_{\boldsymbol{X}}}\{\boldsymbol{\gamma}_{\mu }%
\boldsymbol{X},\boldsymbol{\gamma}_{\sigma}\boldsymbol{X}\}=\boldsymbol{%
\gamma}_{\mu}\underline{\boldsymbol{\sigma}}_{\boldsymbol{X}}^{2}\boldsymbol{%
\gamma}_{\sigma}^{\prime}=\boldsymbol{0}_{\bar{k}_{\mu}\times\bar{k}%
_{\sigma}}\text{,}  \label{uncorr-view}
\end{equation}
the updated expectation (\ref{comp-quad-norm-mu-1}) simplifies as [\ref%
{norm-post-pdf-quad-pol-3}]%
\begin{equation}
\boldsymbol{\bar{\mu}}_{\boldsymbol{X}}=\text{ }\underline{\boldsymbol{\mu}}%
_{\boldsymbol{X}}+\boldsymbol{\gamma}_{\mu}^{\dag}(\boldsymbol{\mu }^{%
\mathit{info}}-\boldsymbol{\gamma}_{\mu}\underline{\boldsymbol{\mu}}_{%
\boldsymbol{X}})+\boldsymbol{\gamma}_{\sigma}^{\dag}(\boldsymbol{\sigma }^{2%
\mathit{info}}(\boldsymbol{\gamma}_{\sigma}\underline{\boldsymbol{\sigma}}_{%
\boldsymbol{X}}^{2}\boldsymbol{\gamma}_{\sigma}^{\prime})^{-1}\boldsymbol{%
\gamma}_{\sigma}\underline{\boldsymbol{\mu}}_{\boldsymbol{X}}-\boldsymbol{%
\gamma}_{\sigma}\underline{\boldsymbol{\mu}}_{\boldsymbol{X}})\text{,}
\label{pseudo-inv-propp-4}
\end{equation}
where the last term on the right hand side is a correction to \cite%
{Meucci08a}.

\section{Numerical results\label{scen-prob-sec}}

We consider base distributions (\ref{prior-distr}) whose analytical
expression is known, possibly up to multiplicative constant term%
\begin{equation}
\underline{f}_{\boldsymbol{X}}(\boldsymbol{x})\propto\underline{g}_{%
\boldsymbol{X}}(\boldsymbol{x})\text{,}  \label{generic-prior}
\end{equation}
for some known analytical function $\underline{g}_{\boldsymbol{X}}$ , which
we call \textquotedblleft numerator\textquotedblright.

Efficient Markov chain Monte Carlo\ (MCMC)\ techniques are available to draw
scenarios from the broad class (\ref{generic-prior}), see \cite%
{ChibGreenberg95} and \cite{Geweke99}%
\begin{equation}
\underline{g}_{\boldsymbol{X}}\quad\overset{\text{MCMC}}{\Rightarrow}\quad\{%
\underline{\boldsymbol{x}}^{(j)},\underline{p}^{(j)}\}_{j=1}^{\bar{\jmath}%
}\approx\underline{f}_{\boldsymbol{X}}\text{.}  \label{base-simul}
\end{equation}
In particular, in our implementations we chose Hamiltonian Monte Carlo
sampling \cite{Chao15}, \cite{Neal11}.

With general inference of expectation type (\ref{view-gen-exp-norm}), the
MRE updated distribution (\ref{exp-family-set}) is an exponential tilt of
the base distribution (\ref{exp-family-set}) and therefore it has again an
analytical expression, up to a constant%
\begin{equation}
\bar{f}_{\boldsymbol{X}}(\boldsymbol{x})\propto\underline{g}_{\boldsymbol{X}%
}(\boldsymbol{x})e^{\boldsymbol{\theta}^{\mathit{info}\prime}\zeta (%
\boldsymbol{x})}\text{,}  \label{generic-post-1}
\end{equation}
for optimal Lagrange multipliers $\boldsymbol{\theta}^{\mathit{info}%
}\equiv(\theta_{1}^{\mathit{info}},\ldots,\theta_{\bar{k}}^{\mathit{info}%
})^{\prime}$ that solve (\ref{lagr-prob}). Therefore, \emph{if} we we can
compute or approximate $\boldsymbol{\theta}^{\mathit{info}}$, we can draw
scenarios from the updated distribution $\bar{f}_{\boldsymbol{X}}$ [\ref%
{hmc-smapling-sec}].

An efficient algorithm to compute an approximate updated distribution $\bar {%
f}_{\boldsymbol{X}}$ and approximate Lagrange multipliers $\hat {\boldsymbol{%
\theta}}^{\mathit{info}}\approx\boldsymbol{\theta}^{\mathit{info}}%
\boldsymbol{\ }$is the discrete MRE \cite{Meucci08c}%
\begin{equation}
\left. 
\begin{array}{l}
\{\underline{\boldsymbol{x}}^{(j)},\underline{p}^{(j)}\}_{j=1}^{\bar{\jmath}%
}\approx\underline{f}_{\boldsymbol{X}} \\ 
\{\zeta,\boldsymbol{\eta}^{\mathit{info}}\}%
\end{array}
\right\} \quad\overset{\text{MRE}}{\Rightarrow}\quad\left\{ 
\begin{array}{l}
\{\underline{\boldsymbol{x}}^{(j)},\bar{p}^{(j)}\}_{j=1}^{\bar{\jmath}%
}\approx\bar{f}_{\boldsymbol{X}} \\ 
\hat{\boldsymbol{\theta}}^{\mathit{info}}\approx\boldsymbol{\theta }^{%
\mathit{info}}\text{.}%
\end{array}
\right.  \label{np-mre-intuition}
\end{equation}
The quality of the approximation $\{\underline{\boldsymbol{x}}^{(j)},\bar {p}%
^{(j)}\}_{j=1}^{\bar{\jmath}}\approx\bar{f}_{\boldsymbol{X}}$ (\ref%
{np-mre-intuition}) can be measured by the discrete relative entropy caused
by the information perturbation, or, equivalently, the exponential of its
negative counterpart, i.e. the effective number of scenarios in \cite%
{MeuEffNumScen12}%
\begin{equation}
\mathit{ens}(\bar{\boldsymbol{p}},\underline{\boldsymbol{p}})\equiv \exp(-%
\mathcal{E}(\bar{\boldsymbol{p}}||\underline{\boldsymbol{p}}))=\exp(\tsum
\nolimits_{j=1}^{\bar{\jmath}}\bar{p}^{(j)}\ln\frac{\bar{p}^{(j)}}{%
\underline{p}^{(\bar{\jmath})}})\text{.}  \label{eff-num-scen}
\end{equation}
The approximation in general is poor for problems of large dimensions $\bar {%
n}$: because the scenarios are the same as the base scenarios, when the
information constraints (\ref{view-gen-exp-norm}) are strongly violated by
the base distribution (\ref{generic-prior}), the curse of dimensionality
forces a few scenarios to carry most of the probability, which amounts to a
too low effective number of scenarios $\mathit{ens}(\bar{\boldsymbol{p}},%
\underline{\boldsymbol{p}})\ll1$ (\ref{eff-num-scen}). Instead, because of
the low dimension of the information constraints (\ref{less-view-mkt}), the
approximate Lagrange multipliers $\hat{\boldsymbol{\theta}}^{\mathit{info}%
}\approx\boldsymbol{\theta}^{\mathit{info}}$ are much more accurate. Here we
show how to exploit this feature to obtain accurate representations of the
updated distribution.

To this purpose, let us write the exact updated numerator (\ref%
{generic-post-1}) as%
\begin{equation}
\underline{g}_{\boldsymbol{X}}(\boldsymbol{x})e^{\boldsymbol{\theta }^{%
\mathit{info}\prime}\zeta(\boldsymbol{x})}=\underline{g}_{\boldsymbol{X}}(%
\boldsymbol{x})e^{\boldsymbol{\hat{\theta}}^{\mathit{info}\prime}\zeta(%
\boldsymbol{x})}\times e^{\Delta\boldsymbol{\theta}^{\prime}\zeta(%
\boldsymbol{x})}\text{,}  \label{decomp-approx}
\end{equation}
which can be interpreted as an MRE tilt as in (\ref{generic-post-1}), but
with a new base%
\begin{equation}
\hat{f}_{\boldsymbol{X}}(\boldsymbol{x})\propto\underline{g}_{\boldsymbol{X}%
}(\boldsymbol{x})e^{\boldsymbol{\hat{\theta}}^{\mathit{info}\prime}\zeta(%
\boldsymbol{x})}\text{;}  \label{new-estim-base}
\end{equation}
and a new Lagrange multipliers%
\begin{equation}
\Delta\boldsymbol{\theta}\equiv\boldsymbol{\theta}^{\mathit{info}}-\hat{%
\boldsymbol{\theta}}^{\mathit{info}}\text{.}  \label{incremental-lagrange}
\end{equation}

As long as the information conditions $\mathcal{C}_{\boldsymbol{X}}$ (\ref%
{view-gen-exp-norm}) are fixed, the true MRE updated distribution $\bar{f}_{%
\boldsymbol{X}}$ (\ref{generic-post-1}) is the same if we replace the
original base $\underline{f}_{\boldsymbol{X}}$ (\ref{generic-prior}) with
the new one $\hat{f}_{\boldsymbol{X}}$ (\ref{new-estim-base}) [\ref%
{composition-appx-1}]%
\begin{equation}
\bar{f}_{\boldsymbol{X}}\equiv\limfunc{argmin}\limits_{f_{\boldsymbol{X}}\in%
\mathcal{C}_{\boldsymbol{X}}}\mathcal{E}(f_{\boldsymbol{X}}\Vert \underline{f%
}_{\boldsymbol{X}})=\limfunc{argmin}\limits_{f_{\boldsymbol{X}}\in\mathcal{C}%
_{\boldsymbol{X}}}\mathcal{E}(f_{\boldsymbol{X}}\Vert\hat{f}_{\boldsymbol{X}%
})\text{.}  \label{same-post}
\end{equation}

Moreover, when the information constraints (\ref{view-gen-exp-norm})
contradicts the base distribution (\ref{generic-prior}), and hence $\mathit{%
ens}(\bar{\boldsymbol{p}},\underline{\boldsymbol{p}})\ll1$, the new base $%
\hat{f}_{\boldsymbol{X}}$ (\ref{new-estim-base}) is \emph{closer} to the
target than the base $\underline{f}_{\boldsymbol{X}}$ (\ref{generic-prior}) 
\begin{equation}
\mathcal{E}(\bar{f}_{\boldsymbol{X}}\Vert\hat{f}_{\boldsymbol{X}})<\mathcal{E%
}(\bar{f}_{\boldsymbol{X}}\Vert\underline{f}_{\boldsymbol{X}})\text{,}
\label{new-base-closer}
\end{equation}
because the numerical MRE\ multipliers (\ref{np-mre-intuition}) are close to
the true ones $\hat{\boldsymbol{\theta}}^{\mathit{info}}\approx \boldsymbol{%
\theta}^{\mathit{info}}$.

Hence, we can generate \emph{new} scenarios from the updated base (\ref%
{new-estim-base})%
\begin{equation}
\underline{g}_{\boldsymbol{X}}(\boldsymbol{x})e^{\boldsymbol{\hat{\theta}}^{%
\mathit{info}\prime}\zeta(\boldsymbol{x})}\quad\overset{\text{MCMC}}{%
\Rightarrow}\quad\{\bar{\boldsymbol{x}}^{(j)},\underline{p}^{(j)}\}_{j=1}^{%
\bar{\jmath}}\approx\hat{f}_{\boldsymbol{X}}\text{,}  \label{simul-hmc}
\end{equation}
and use the simulation output as input for the discrete MRE algorithm (\ref%
{np-mre-intuition})\ to obtain new multipliers $\Delta\hat {\boldsymbol{%
\theta}}$ and new probabilities $\{\bar{p}^{(j)}\}_{j=1}^{\bar{\jmath}}$%
\begin{equation}
\left. 
\begin{array}{l}
\{\bar{\boldsymbol{x}}^{(j)},\underline{p}^{(j)}\}_{j=1}^{\bar{\jmath}%
}\approx\hat{f}_{\boldsymbol{X}} \\ 
\{\zeta,\boldsymbol{\eta}^{\mathit{info}}\}%
\end{array}
\right\} \quad\overset{\text{MRE}}{\Rightarrow}\quad\left\{ 
\begin{array}{l}
\{\bar{\boldsymbol{x}}^{(j)},\bar{p}^{(j)}\}_{j=1}^{\bar{\jmath}}\approx 
\bar{f}_{\boldsymbol{X}} \\ 
\Delta\hat{\boldsymbol{\theta}}\approx\boldsymbol{\theta}^{\mathit{info}}-%
\hat{\boldsymbol{\theta}}^{\mathit{info}}%
\end{array}
\right. \text{.}  \label{mre-new}
\end{equation}

The quality of the approximation $\{\bar{\boldsymbol{x}}^{(j)},\bar{p}%
^{(j)}\}_{j=1}^{\bar{\jmath}}\approx\bar{f}_{\boldsymbol{X}}$ (\ref{mre-new}%
) is better than the original output $\{\underline{\boldsymbol{x}}^{(j)},%
\bar {p}^{(j)}\}_{j=1}^{\bar{\jmath}}\approx\bar{f}_{\boldsymbol{X}}$ (\ref%
{np-mre-intuition}), because here the starting point $\hat{f}_{\boldsymbol{X}%
}$ is closer to the MRE updated distribution $\bar {f}_{\boldsymbol{X}}$ (%
\ref{new-base-closer}) and thus the curse of dimensionality is mitigated.
Furthermore, the new output $\{\bar {\boldsymbol{x}}^{(j)},\bar{p}%
^{(j)}\}_{j=1}^{\bar{\jmath}}$ respects the inference constraints (\ref%
{view-gen-exp-norm}) exactly%
\begin{equation}
\tsum \nolimits_{j=1}^{\bar{\jmath}}\bar{p}^{(j)}\zeta(\bar{\boldsymbol{x}}%
^{\left( j\right) })=\boldsymbol{\eta }^{\mathit{info}}\text{,}
\label{view-satis-non-param}
\end{equation}
unlike the simulation input $\{\bar{\boldsymbol{x}}^{(j)},\underline{p}%
^{(j)}\}_{j=1}^{\bar{\jmath}}$ (\ref{simul-hmc}).

Then we can update the Lagrange multipliers 
\begin{equation}
\hat{\boldsymbol{\theta}}^{\mathit{info}}\leftarrow\hat{\boldsymbol{\theta}}%
^{\mathit{info}}+\Delta\hat{\boldsymbol{\theta}}\text{,}
\label{update-evidence}
\end{equation}
and iterate (\ref{simul-hmc})-(\ref{mre-new}). Convergence in the above
routine occurs when the effective number of scenarios (\ref{eff-num-scen})
falls above a given threshold%
\begin{equation}
\mathit{ens}(\bar{\boldsymbol{p}},\underline{\boldsymbol{p}})>1-\delta\text{,%
}  \label{check-mre}
\end{equation}
where $0<\delta\ll1$.

We summarize the iterative MRE in the following table.

\begin{table}[H]\begin{center}
\renewcommand{\arraystretch}{1.2}%
\begin{tabular}{ll}
\multicolumn{2}{c}{$(\{\bar{\boldsymbol{x}}^{(j)},\bar{p}^{(j)}\}_{j=1}^{%
\bar{\jmath}},\bar{g}_{\boldsymbol{X}})=\mathit{MRE.Iterative}(\zeta,%
\boldsymbol{\eta}^{\mathit{info}},\underline{g}_{\boldsymbol{X}},\bar{\jmath}%
,\delta)$} \\ \hline
0. Initialize numerator & $\bar{g}_{\boldsymbol{X}}\leftarrow\underline{g}_{%
\boldsymbol{X}}$ \\ 
1. Generate new scenarios & $\{\bar{\boldsymbol{x}}^{(j)},\underline{p}%
^{(j)}\}_{j=1}^{\bar{\jmath}}\overset{\text{MCMC}}{\Leftarrow}(\bar {g}_{%
\boldsymbol{X}},\bar{\jmath})\text{ (\ref{base-simul})}$ \\ 
2. Perform discrete MRE & $(\Delta\hat{\boldsymbol{\theta}},\{\bar{p}%
^{(j)}\}_{j=1}^{\bar{\jmath}})\overset{\text{MRE}}{\Leftarrow}(\{\bar {%
\boldsymbol{x}}^{(j)},\underline{p}^{(j)}\}_{j=1}^{\bar{\jmath}},\zeta,%
\boldsymbol{\eta}^{\mathit{info}})$ $\text{(\ref{mre-new})}$ \\ 
3. Update Lagrange multipliers & $\hat{\boldsymbol{\theta}}^{\mathit{info}%
}\leftarrow\hat{\boldsymbol{\theta}}^{\mathit{info}}+\Delta\hat {\boldsymbol{%
\theta}}\text{ (\ref{view-satis-non-param})}$ \\ 
4. Update numerator & $\bar{g}_{\boldsymbol{X}}(\cdot)\leftarrow \underline{g%
}_{\boldsymbol{X}}(\cdot)e^{\hat{\boldsymbol{\theta}}^{\mathit{info}%
\prime}\zeta(\cdot)}$ (\ref{new-estim-base}) \\ 
5. Check convergence & $\mathit{ens}(\bar{\boldsymbol{p}},\underline{%
\boldsymbol{p}})>1-\delta$ (\ref{check-mre}) \\ 
\multicolumn{2}{l}{6. If convergence, output $(\{\bar{\boldsymbol{x}}^{(j)},%
\bar{p}^{(j)}\}_{j=1}^{\bar{\jmath}},\bar{g}_{\boldsymbol{X}})$; else go to 1
} \\ \hline
\end{tabular}
\caption{\label{iter-mre-routine}Iterative MRE algorithm.}%
\end{center}\end{table}\renewcommand{\arraystretch}{1}%

\section{A case study\label{sec-application}}

We consider $\bar{n}\equiv7$ target variables $\boldsymbol{X}\equiv\left(
X_{1},X_{2},\ldots,X_{7}\right) ^{\prime}$ with normal base distribution (%
\ref{norm-exp-form})%
\begin{equation}
\boldsymbol{X}\sim\mathit{N}(\underline{\boldsymbol{\mu}}_{\boldsymbol{X}},%
\underline{\boldsymbol{\sigma}}_{\boldsymbol{X}}^{2})\text{,}
\label{x-norm-base}
\end{equation}
and homogeneous expectations, standard deviations%
\begin{equation}
\underline{\boldsymbol{\mu}}_{\boldsymbol{X}}=\left( 
\begin{smallmatrix}
10\% \\ 
10\% \\ 
10\% \\ 
10\% \\ 
10\% \\ 
10\% \\ 
10\%%
\end{smallmatrix}
\right) \text{,}\quad\mathit{diag}(\underline{\boldsymbol{\sigma}}_{%
\boldsymbol{X}}^{2})=\left( 
\begin{smallmatrix}
20\% \\ 
20\% \\ 
20\% \\ 
20\% \\ 
20\% \\ 
20\% \\ 
20\%%
\end{smallmatrix}
\right) \text{;}  \label{x-norm-mu-sig}
\end{equation}
and homogeneous correlations%
\begin{equation}
\mathit{corr}(\underline{\boldsymbol{\sigma}}_{\boldsymbol{X}}^{2})=\left( 
\begin{smallmatrix}
100\% & 70\% & 70\% & 70\% & 70\% & 70\% & 70\% \\ 
\cdot & 100\% & 70\% & 70\% & 70\% & 70\% & 70\% \\ 
\cdot & \cdot & 100\% & 70\% & 70\% & 70\% & 70\% \\ 
\cdot & \cdot & \cdot & 100\% & 70\% & 70\% & 70\% \\ 
\cdot & \cdot & \cdot & \cdot & 100\% & 70\% & 70\% \\ 
\cdot & \cdot & \cdot & \cdot & \cdot & 100\% & 70\% \\ 
\cdot & \cdot & \cdot & \cdot & \cdot & \cdot & 100\%%
\end{smallmatrix}
\right) \text{.}  \label{x-norm-mu-sig2}
\end{equation}

Then we consider information constraints (\ref{part-view-pos}) as follows%
\begin{equation}
f_{\boldsymbol{X}}\in\mathcal{C}_{\boldsymbol{X}}:\qquad\left\{ 
\begin{array}{l}
\mathbb{E}^{f_{\boldsymbol{X}}}\{X_{3}\}=35\% \\ 
\mathbb{C}r^{f_{\boldsymbol{X}}}\{X_{1},X_{2}\}=-80\%\text{.}%
\end{array}
\right.  \label{x-view-cor}
\end{equation}%
\FRAME{fhFU}{342pt}{234.75pt}{0pt}{\Qcb{MRE updated distribution under
normal base (\protect\ref{x-norm-base}) and inference constraints (\protect
\ref{x-view-cor}). In green the location-dispersion ellipsoid and
simulations from the base distribution. In orange and red the
location-dispersion ellipsoids stemming from the first and second step
simulations via iterative approach (Table \protect\ref{iter-mre-routine}),
respectively. In black the location-dispersion ellispoid of the analytical
solution and the third-step simulations.}}{\Qlb{fig-exampl-1}}{%
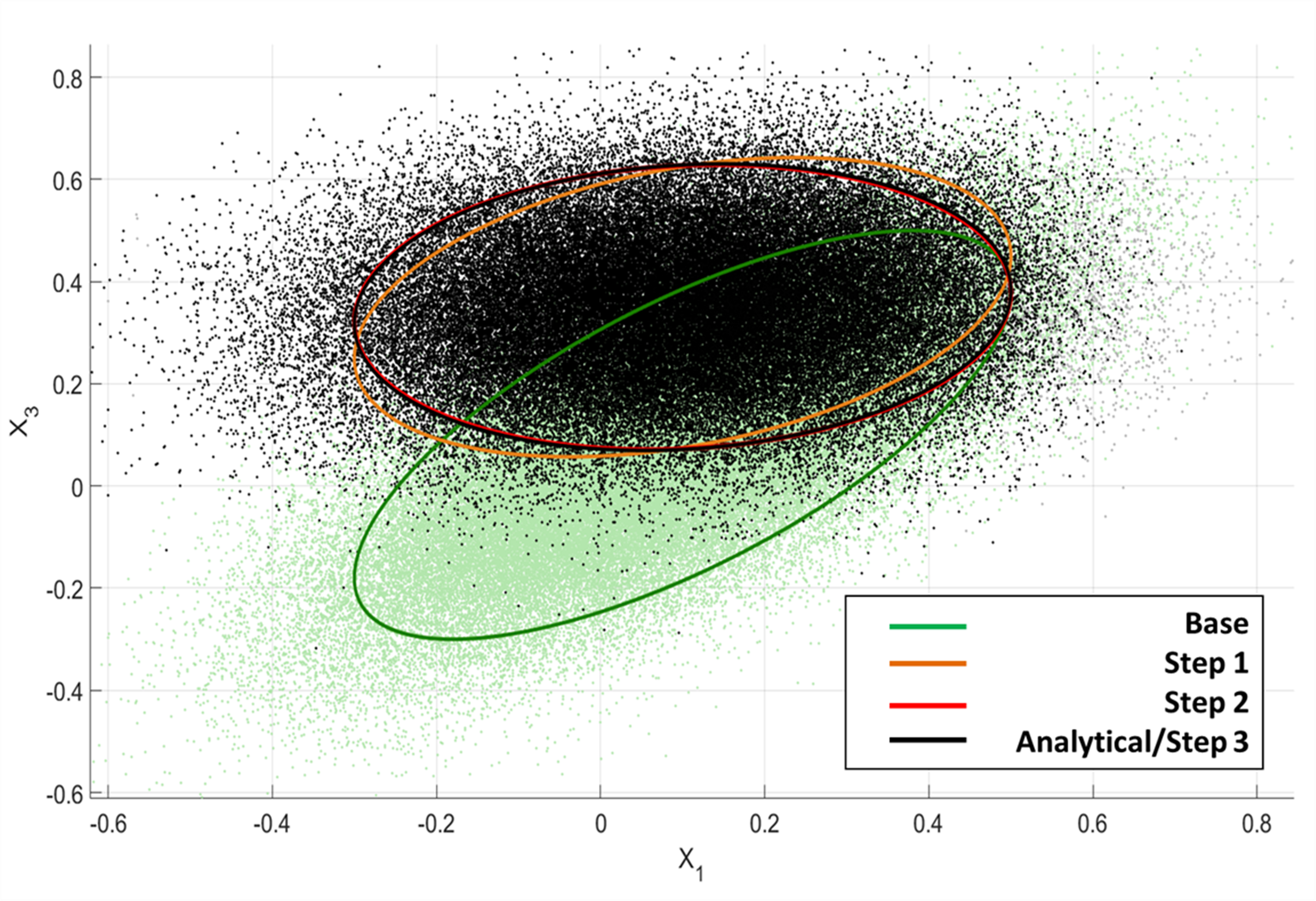}{\special{ language "Scientific Word"; type "GRAPHIC";
maintain-aspect-ratio TRUE; display "USEDEF"; valid_file "F"; width 342pt;
height 234.75pt; depth 0pt; original-width 9.9799in; original-height
6.832in; cropleft "0"; croptop "1"; cropright "1"; cropbottom "0"; filename
'fig-exampl-1.png';file-properties "XNPEU";}} Also, we assume that the
constraints on correlations (\ref{x-view-cor}) do not alter the respective
first and second moments of the variables $X_{1},X_{2}$, so that we can
rewrite the information (\ref{x-view-cor}) as expectation conditions (\ref%
{view-gen-exp-norm})%
\begin{equation}
f_{\boldsymbol{X}}\in\mathcal{C}_{\boldsymbol{X}}:\qquad\left\{ 
\begin{array}{l}
\mathbb{E}^{f_{\boldsymbol{X}}}\{X_{1}\}=\mathbb{E}^{f_{\boldsymbol{X}%
}}\{X_{2}\}=10\% \\ 
\mathbb{E}^{f_{\boldsymbol{X}}}\{X_{3}\}=35\% \\ 
\mathbb{E}^{f_{\boldsymbol{X}}}\{X_{1}^{2}\}=\mathbb{E}^{f_{\boldsymbol{X}%
}}\{X_{2}^{2}\}=(20\%)^{2}+(10\%)^{2} \\ 
\mathbb{E}^{f_{\boldsymbol{X}}}\{X_{1}X_{2}\}=-80\%\times(20\%)^{2}+(10%
\%)^{2}\text{.}%
\end{array}
\right.  \label{x-views}
\end{equation}

We simulate $\bar{\jmath}\equiv100,000$ scenarios with uniform probabilities
(\ref{base-simul}) from the normal base distribution (\ref{x-norm-base}).
Then, from the base scenarios and the information (\ref{x-views}) we compute
the MRE updated distribution (\ref{generic-post-1}) using the iterative
numerical routine (\ref{iter-mre-routine}). The routine reaches convergence
in three steps with a threshold $\delta\equiv0.01$ (\ref{check-mre}).

Equivalently, we can express the information (\ref{x-views})\ as constraints
on linear combinations of expectations and covariances as in (\ref%
{view-exp-cov-norm}), where:

\begin{itemize}
\item $\boldsymbol{\gamma}_{\mu}$ is a $3\times7$ matrix as follows%
\begin{equation}
\boldsymbol{\gamma}_{\mu}\equiv\left( 
\begin{smallmatrix}
1 & 0 & 0 & 0 & 0 & 0 & 0 \\ 
0 & 1 & 0 & 0 & 0 & 0 & 0 \\ 
0 & 0 & 1 & 0 & 0 & 0 & 0%
\end{smallmatrix}
\right) \text{;}  \label{x-views-2}
\end{equation}

\item $\boldsymbol{\mu}^{\mathit{info}}$ is a $3\times1$ vector as follows%
\begin{equation}
\boldsymbol{\mu}^{\mathit{info}}\equiv\left( 
\begin{smallmatrix}
10\% \\ 
10\% \\ 
35\%%
\end{smallmatrix}
\right) \text{;}  \label{x-views-3}
\end{equation}

\item $\boldsymbol{\gamma}_{\sigma}$ is a $2\times7$ matrix as follows 
\begin{equation}
\boldsymbol{\gamma}_{\sigma}\equiv\left( 
\begin{smallmatrix}
1 & 0 & 0 & 0 & 0 & 0 & 0 \\ 
0 & 1 & 0 & 0 & 0 & 0 & 0%
\end{smallmatrix}
\right) \text{;}  \label{x-views-4}
\end{equation}

\item $\boldsymbol{\sigma}^{2\mathit{info}}$ is a $2\times2$ matrix as
follows%
\begin{equation}
\boldsymbol{\sigma}^{2\mathit{info}}\equiv\left( 
\begin{smallmatrix}
(20\%)^{2} & -80\%\times(20\%)^{2} \\ 
-80\%\times(20\%)^{2} & (20\%)^{2}%
\end{smallmatrix}
\right) \text{.}  \label{x-views-5}
\end{equation}
\end{itemize}

Then, from the base normal distribution (\ref{x-norm-base}) and the
information (\ref{x-views}), we compute analytically the MRE updated
distribution (\ref{mre-posterior}), which is normal (\ref{pos-exp-nomr-1})%
\begin{equation}
\boldsymbol{X}\sim\mathit{N}(\boldsymbol{\bar{\mu}}_{\boldsymbol{X}},\bar{%
\boldsymbol{\sigma}}_{\boldsymbol{X}}^{2})\text{,}  \label{x-update}
\end{equation}
where the updated expectations (\ref{comp-quad-norm-mu-1}) and standard
deviations (\ref{comp-quad-norm-sig-1}) read%
\begin{equation}
\boldsymbol{\bar{\mu}}_{\boldsymbol{X}}=\left( 
\begin{smallmatrix}
10\% \\ 
10\% \\ 
35\% \\ 
17.29\% \\ 
17.29\% \\ 
17.29\% \\ 
17.29\%%
\end{smallmatrix}
\right) \text{,}\quad\mathit{diag}(\boldsymbol{\bar{\sigma}}_{\boldsymbol{X}%
}^{2})=\left( 
\begin{smallmatrix}
20\% \\ 
20\% \\ 
14.02\% \\ 
14.02\% \\ 
14.02\% \\ 
14.02\% \\ 
14.02\%%
\end{smallmatrix}
\right) \text{;}  \label{x-update-mu-sig2}
\end{equation}
and the updated correlations (\ref{comp-quad-norm-sig-1}) read%
\begin{equation}
\mathit{corr}(\boldsymbol{\bar{\sigma}}_{\boldsymbol{X}}^{2})=\left( 
\begin{smallmatrix}
100\% & -80\% & 11.75\% & 11.75\% & 11.75\% & 11.75\% & 11.75\% \\ 
\cdot & 100\% & 11.75\% & 11.75\% & 11.75\% & 11.75\% & 11.75\% \\ 
\cdot & \cdot & 100\% & 38.94\% & 38.94\% & 38.94\% & 38.94\% \\ 
\cdot & \cdot & \cdot & 100\% & 38.94\% & 38.94\% & 38.94\% \\ 
\cdot & \cdot & \cdot & \cdot & 100\% & 38.94\% & 38.94\% \\ 
\cdot & \cdot & \cdot & \cdot & \cdot & 100\% & 38.94\% \\ 
\cdot & \cdot & \cdot & \cdot & \cdot & \cdot & 100\%%
\end{smallmatrix}
\right) \text{.}  \label{update-corr}
\end{equation}

In Figure \ref{fig-exampl-1} we report the results of numerical and
analytical approaches, and in the following Table \ref%
{iter-non-par-ens-comparson} we summarize the errors between the respective
statistics.

\begin{table}[H]\begin{center}
\renewcommand{\arraystretch}{1.2}%
$%
\begin{tabular}{c|ccc}
& $\mathit{ens}(\bar{\boldsymbol{p}},\underline{\boldsymbol{p}})$ & $||\hat{%
\boldsymbol{\mu}}_{\boldsymbol{X}}-\boldsymbol{\bar{\mu}}_{\boldsymbol{X}}||$
& $||\hat{\boldsymbol{\sigma}}_{\boldsymbol{X}}^{2}-\boldsymbol{\bar{\sigma}}%
_{\boldsymbol{X}}^{2}||_{F}$ \\ \hline
Step 1 & $2.24\%$ & $1.04\times10^{-2}$ & $2.57\times10^{-2}$ \\ 
Step 2 & $81.32\%$ & $0.15\times10^{-2}$ & $1.15\times10^{-3}$ \\ 
Step 3 & $99.97\%$ & $0.11\times10^{-2}$ & $7.11\times10^{-4}$%
\end{tabular}
\ $%
\caption{\label{iter-non-par-ens-comparson}Iterative MRE: effective
number of scenarios and errors.}%
\end{center}\end{table}\renewcommand{\arraystretch}{1}%

\section{Conclusions\label{conclusions-1}}

In this article we showed how to solve analytically and numerically the MRE
problem under exponential-family base distributions and partial information
constraints of expectation type as in (\ref{quad-view-func-norm}).

Under normal base distributions, we computed analytically the MRE solution (%
\ref{pos-exp-nomr-1}) and fixed the formulation of the updated expectation
originally proposed by \cite{Meucci08a}.

Under more general base distributions, we showed how to compute numerically
the MRE solution via iterative Hamiltonian Monte Carlo simulations (Table %
\ref{iter-mre-routine}) yielding a better approximation of the updated
distribution than the original scenario-based algorithm\ in \cite{Meucci08a}.

\cleardoublepage%

\bibliographystyle{apalike}
\bibliography{finance}
\newpage

\appendix

\section{Appendix}

Here we discuss some technical results of Sections \ref%
{sec-normal-assumption} and \ref{scen-prob-sec}.

\subsection{MRE with exponential-family base\label{gen-post-pdf-appendix}}

Consider a base distribution $\underline{f}_{\boldsymbol{X}}$ (\ref%
{prior-distr}) in the exponential family class $\mathit{Exp}(\underline{%
\boldsymbol{\theta}}_{\boldsymbol{X}},\tau,h,\mathcal{X})$ as in (\ref%
{exp-fam-base}), where $\underline{\boldsymbol{\theta}}_{\boldsymbol{X}%
}\in\Theta$, and hence with the following pdf%
\begin{equation}
\underline{f}_{\boldsymbol{X}}(\boldsymbol{x})=h(\boldsymbol{x})\exp (%
\underline{\boldsymbol{\theta}}_{\boldsymbol{X}}^{\prime}\tau(\boldsymbol{x}%
)-\psi_{h,\tau}(\underline{\boldsymbol{\theta}}_{\boldsymbol{X}}))\text{,}
\label{exp-fam-base-pdf}
\end{equation}
where $\psi_{h,\tau}$ denotes the log-partition function as in (\ref%
{soidfodngf})%
\begin{equation}
\psi_{h,\tau}(\boldsymbol{\theta})\equiv\ln\int_{\mathbb{R}^{\bar{n}}}e^{%
\boldsymbol{\theta}^{\prime}\tau(\boldsymbol{x})}\underline{f}_{\boldsymbol{X%
}}\left( \boldsymbol{x}\right) d\boldsymbol{x}\text{.}  \label{gen-log-part}
\end{equation}

Then the updated distribution $\bar{f}_{\boldsymbol{X}}$ (\ref%
{exp-family-set-1}) reads%
\begin{align}
\bar{f}_{\boldsymbol{X}}(\boldsymbol{x}) & =\underline{f}_{\boldsymbol{X}}(%
\boldsymbol{x})\exp(\boldsymbol{\theta}^{\mathit{info}\prime}\zeta (%
\boldsymbol{x})-\psi_{\underline{f}_{\boldsymbol{X}},\zeta}(\boldsymbol{%
\theta}^{\mathit{info}}))  \notag \\
& =h(\boldsymbol{x})\exp(\underline{\boldsymbol{\theta}}_{\boldsymbol{X}%
}^{\prime}\tau(\boldsymbol{x})-\psi_{h,\tau}(\underline{\boldsymbol{\theta}}%
_{\boldsymbol{X}}))\times\exp(\boldsymbol{\theta}^{\mathit{info}\prime }%
\boldsymbol{\gamma}\tau(\boldsymbol{x})-\psi_{\underline{f}_{\boldsymbol{X}%
},\zeta}(\boldsymbol{\theta}^{\mathit{info}}))  \notag \\
& =h(\boldsymbol{x})\exp(\boldsymbol{\bar{\theta}}_{\boldsymbol{X}}^{\prime
}\tau(\boldsymbol{x})-\psi_{h,\tau}(\underline{\boldsymbol{\theta}}_{%
\boldsymbol{X}})-\psi_{\underline{f}_{\boldsymbol{X}},\zeta}(\boldsymbol{%
\theta}^{\mathit{info}}))\text{,}  \label{gen-log-part-1}
\end{align}
where in the second row we used the linearity of the inference functions $%
\zeta$ with respect to the sufficient statistics\textbf{\ }$\tau$ as in (\ref%
{quad-view-func-norm}); and where we defined 
\begin{equation}
\boldsymbol{\bar{\theta}}_{\boldsymbol{X}}\equiv\underline{\boldsymbol{%
\theta }}_{\boldsymbol{X}}+\boldsymbol{\gamma}^{\prime}\boldsymbol{\theta }^{%
\mathit{info}}\text{,}  \label{new-can-coord-lin-upd}
\end{equation}
as in (\ref{post-can-param-norm}). Then, as long as $\boldsymbol{\bar{\theta}%
}_{\boldsymbol{X}}\in\Theta$, the log-partition functions (\ref{gen-log-part}%
) satisfy%
\begin{equation}
\psi_{h,\tau}(\boldsymbol{\bar{\theta}}_{\boldsymbol{X}})=\psi_{h,\tau }(%
\underline{\boldsymbol{\theta}}_{\boldsymbol{X}})+\psi_{\underline{f}_{%
\boldsymbol{X}},\zeta}(\boldsymbol{\theta}^{\mathit{info}})\text{,}
\label{gen-log-part-2}
\end{equation}
which implies our desired result (\ref{exp-fam-conj-update}).

\subsection{MRE update with normal base and information on non-central
moments \label{norm-post-pdf-appendix}}

The pdf of the normal base distribution $\underline{f}_{\boldsymbol{X}}$ as
in (\ref{norm-exp-form}) can be written in canonical form within the
exponential family class $\mathit{Exp}(\underline{\boldsymbol{\theta}}_{%
\boldsymbol{X}}^{\mathit{N}},\tau^{\mathit{N}},h^{\mathit{N}},\mathbb{R}^{%
\bar{n}})$ (\ref{norm-exp-form}) as follows%
\begin{equation}
\underline{f}_{\boldsymbol{X}}(\boldsymbol{x})=(2\pi)^{-\frac{\bar{n}}{2}%
}\exp(\underline{\boldsymbol{\theta}}_{\boldsymbol{X};\mu}^{\mathit{N}\prime
}\boldsymbol{x}+\mathit{vec}(\underline{\boldsymbol{\theta}}_{\boldsymbol{X}%
;\sigma,\sigma}^{\mathit{N}})^{\prime}\mathit{vec}(\boldsymbol{xx}^{\prime
})-\psi^{\mathit{N}}(\underline{\boldsymbol{\theta}}_{\boldsymbol{X}}^{%
\mathit{N}}))\text{,}  \label{norm-pdf-can}
\end{equation}
where $\underline{\boldsymbol{\theta}}_{\boldsymbol{X};\mu}^{\mathit{N}}$
and $\underline{\boldsymbol{\theta}}_{\boldsymbol{X};\sigma,\sigma}^{\mathit{%
N}}$ identify the base canonical coordinates $\underline{\boldsymbol{\theta}}%
_{\boldsymbol{X}}^{\mathit{N}}$ (\ref{norm-can-cord-pri}); and where
log-partition function (\ref{gen-log-part}), with respect to the reference
measure $h^{\mathit{N}}(\boldsymbol{x})\equiv(2\pi)^{-\bar{n}/2}$ and
sufficient statistics $\tau^{\mathit{N}}$ (\ref{norm-suff-stat-pri}) reads%
\begin{equation}
\psi^{\mathit{N}}(\underline{\boldsymbol{\theta}}_{\boldsymbol{X}}^{\mathit{N%
}})\equiv\psi_{h^{\mathit{N}},\tau^{\mathit{N}}}(\underline{\boldsymbol{%
\theta}}_{\boldsymbol{X}}^{\mathit{N}})=-\tfrac{1}{4}\underline{\boldsymbol{%
\theta}}_{\boldsymbol{X};\mu}^{\mathit{N}\prime }(\underline{\boldsymbol{%
\theta}}_{\boldsymbol{X};\sigma,\sigma}^{\mathit{N}})^{-1}\underline{%
\boldsymbol{\theta}}_{\boldsymbol{X};\mu}^{\mathit{N}}-\tfrac{1}{2}\ln\det(-2%
\underline{\boldsymbol{\theta}}_{\boldsymbol{X};\sigma,\sigma}^{\mathit{N}})%
\text{,}  \label{norm-log-part-func}
\end{equation}
e.g. \cite{AmariNagaoka00} and \cite{AmariShun16}.

Let us consider information constraints on the first two non-central moments
of the target variables 
\begin{equation}
f_{\boldsymbol{X}}\in\mathcal{C}_{\boldsymbol{X}}:\qquad\left\{ 
\begin{array}{l}
\mathbb{E}^{f_{\boldsymbol{X}}}\{\boldsymbol{\gamma}_{\mu}\boldsymbol{X}\}=%
\boldsymbol{\eta}_{\mu}^{\mathit{info}} \\ 
\mathbb{E}^{f_{\boldsymbol{X}}}\{\boldsymbol{\gamma}_{\sigma}\boldsymbol{XX}%
^{\prime}\boldsymbol{\gamma}_{\sigma}^{\prime}\}=\boldsymbol{\eta}%
_{\sigma,\sigma}^{\mathit{info}}\text{,}%
\end{array}
\right.  \label{eq-mom-cond-norm}
\end{equation}
where $\boldsymbol{\eta}_{\mu}^{\mathit{info}}$ is a $\bar{k}_{\mu}\times1$
vector and $\boldsymbol{\gamma}_{\mu}$ is a $\bar{k}_{\mu}\times\bar{n}$
matrix; $\boldsymbol{\eta}_{\sigma,\sigma}^{\mathit{info}}$ is a $\bar {k}%
_{\sigma}\times\bar{k}_{\sigma}$ symmetric matrix and $\boldsymbol{\gamma }%
_{\sigma}$ is a $\bar{k}_{\sigma}\times\bar{n}$ matrix.

Using matrix algebra (see e.g. \cite{magnus79}), we can express the
information constraints (\ref{eq-mom-cond-norm}) as generalized expectation
conditions (\ref{view-gen-exp-norm}) on linear transformations of the normal
sufficient statistics\textbf{\ }$\tau^{\mathit{N}}$ (\ref{norm-suff-stat-pri}%
)%
\begin{equation}
f_{\boldsymbol{X}}\in\mathcal{C}_{\boldsymbol{X}}:\qquad\mathbb{E}^{f_{%
\boldsymbol{X}}}\{\boldsymbol{\gamma}\tau^{\mathit{N}}(\boldsymbol{x})\}=%
\boldsymbol{\eta}^{\mathit{info}}\text{,}  \label{eq-mom-cond-norm-1}
\end{equation}
where:

\begin{itemize}
\item $\boldsymbol{\gamma}$ is the $\bar{k}\times(\bar{n}+\bar{n}^{2})$
matrix defined as follows%
\begin{equation}
\boldsymbol{\gamma}\equiv\left( 
\begin{matrix}
\boldsymbol{\gamma}_{\mu} & \boldsymbol{0}_{\bar{k}_{\mu}\times\bar{n}^{2}}
\\ 
\boldsymbol{0}_{\bar{k}_{\sigma}^{2}\times\bar{n}} & \boldsymbol{\gamma }%
_{\sigma}\otimes\boldsymbol{\gamma}_{\sigma}%
\end{matrix}
\right) \text{;}  \label{zeta-norm-view-mat}
\end{equation}

\item $\boldsymbol{\eta}^{\mathit{info}}$ is the $\bar{k}\times1$ vector
defined as follows%
\begin{equation}
\boldsymbol{\eta}^{\mathit{info}}\equiv\left( 
\begin{array}{c}
\boldsymbol{\eta}_{\mu}^{\mathit{info}} \\ 
\mathit{vec}(\boldsymbol{\eta}_{\sigma,\sigma}^{\mathit{info}})%
\end{array}
\right) \text{.}  \label{vector-eta-norm-view}
\end{equation}
\end{itemize}

Then, according to (\ref{exp-fam-conj-update}), the ensuing updated
distribution (\ref{exp-family-set}) must be in the same exponential family
class of the base $\underline{f}_{\boldsymbol{X}}$ (\ref{norm-exp-form}),
and hence normal in turn (\ref{norm-exp-form})%
\begin{equation}
\bar{f}_{\boldsymbol{X}}\quad\Leftrightarrow\quad\mathit{N}(\bar {%
\boldsymbol{\mu}}_{\boldsymbol{X}},\boldsymbol{\bar{\sigma}}_{\boldsymbol{X}%
}^{2})\quad\Leftrightarrow\mathit{Exp}(\boldsymbol{\bar{\theta}}_{%
\boldsymbol{X}},\tau^{\mathit{N}},h^{\mathit{N}},\mathbb{R}^{\bar{n}})\text{,%
}  \label{up-norm-exp-form}
\end{equation}
where the $\bar{n}\times1$ vector updated canonical coordinates $\bar {%
\boldsymbol{\theta}}_{\boldsymbol{X}}$ read as in (\ref{post-can-param-norm}%
) and where $\boldsymbol{\theta}^{\mathit{info}}$ is the $\bar{k}\times1$
vector of optimal Lagrange multipliers (\ref{lagr-prob}), which we arrange
as follows%
\begin{equation}
\boldsymbol{\theta}^{\mathit{info}}\equiv\left( 
\begin{array}{c}
\boldsymbol{\theta}_{\mu}^{\mathit{info}} \\ 
\mathit{vec}(\boldsymbol{\theta}_{\sigma,\sigma}^{\mathit{info}})%
\end{array}
\right) \text{.}  \label{post-lagr-mult-norm}
\end{equation}
Moreover, the updated expectation in (\ref{pos-exp-nomr-1}) follows from the
updated canonical coordinates $\boldsymbol{\bar{\theta}}_{\boldsymbol{X}}^{%
\mathit{N}}$ and reads%
\begin{equation}
\boldsymbol{\bar{\mu}}_{\boldsymbol{X}}=-\frac{1}{2}(\boldsymbol{\bar{\theta}%
}_{\boldsymbol{X};\sigma,\sigma}^{\mathit{N}})^{-1}\bar{\boldsymbol{\theta}}%
_{\boldsymbol{X};\mu}^{\mathit{N}}\text{,}  \label{pos-exp-nomr-2}
\end{equation}
and similar for the updated covariance in (\ref{pos-exp-nomr-1})%
\begin{equation}
\boldsymbol{\bar{\sigma}}_{\boldsymbol{X}}^{2}=-\frac{1}{2}(\bar {%
\boldsymbol{\theta}}_{\boldsymbol{X};\sigma,\sigma}^{\mathit{N}})^{-1}\text{.%
}  \label{pos-exp-nomr-2-sig}
\end{equation}

In particular, using the linearity of the canonical coordinates $\bar {%
\boldsymbol{\theta}}_{\boldsymbol{X}}^{\mathit{N}}$ (\ref%
{post-can-param-norm}), we can write the updated expectation (\ref%
{pos-exp-nomr-2}) as follows%
\begin{equation}
\boldsymbol{\bar{\mu}}_{\boldsymbol{X}}=\boldsymbol{\bar{\sigma}}_{%
\boldsymbol{X}}^{2}(\underline{\boldsymbol{\theta}}_{\boldsymbol{X};\mu }^{%
\mathit{N}}+\boldsymbol{\gamma}_{\mu}^{\prime}\boldsymbol{\theta}_{\mu }^{%
\mathit{info}})\text{;}  \label{mu-yet}
\end{equation}
and the updated covariance (\ref{pos-exp-nomr-2-sig}) as follows%
\begin{equation}
\boldsymbol{\bar{\sigma}}_{\boldsymbol{X}}^{2}=-\frac{1}{2}(\underline{%
\boldsymbol{\theta}}_{\boldsymbol{X};\sigma,\sigma}^{\mathit{N}}+\boldsymbol{%
\gamma}_{\sigma}^{\prime}\boldsymbol{\theta}_{\sigma,\sigma }^{\mathit{info}}%
\boldsymbol{\gamma}_{\sigma})^{-1}\text{.}  \label{sig2-yet}
\end{equation}

Now, since the updated distribution $\bar{f}_{\boldsymbol{X}}$ (\ref%
{up-norm-exp-form}) must satisfy the information constraints (\ref%
{eq-mom-cond-norm}), then we must have the following equations for the first
moments%
\begin{align}
\mathbb{E}^{\bar{f}_{\boldsymbol{X}}}\{\boldsymbol{\gamma}_{\mu}\boldsymbol{X%
}\} & =\boldsymbol{\gamma}_{\mu}\boldsymbol{\bar{\mu}}_{\boldsymbol{X}}
\label{grad-pol-quad-view-mu} \\
& =\boldsymbol{\gamma}_{\mu}\boldsymbol{\bar{\sigma}}_{\boldsymbol{X}}^{2}(%
\underline{\boldsymbol{\theta}}_{\boldsymbol{X};\mu}^{\mathit{N}}+%
\boldsymbol{\gamma}_{\mu}^{\prime}\boldsymbol{\theta}_{\mu}^{\mathit{info}})=%
\boldsymbol{\eta}_{\mu}^{\mathit{info}}\text{,}  \notag
\end{align}
and second moments%
\begin{align}
\mathbb{E}^{\bar{f}_{\boldsymbol{X}}}\{\boldsymbol{\gamma}_{\sigma }%
\boldsymbol{XX}^{\prime}\boldsymbol{\gamma}_{\sigma}^{\prime}\} & =%
\boldsymbol{\gamma}_{\sigma}(\boldsymbol{\bar{\sigma}}_{\boldsymbol{X}}^{2}+%
\boldsymbol{\bar{\mu}}_{\boldsymbol{X}}\boldsymbol{\bar{\mu}}_{\boldsymbol{X}%
}^{\prime})\boldsymbol{\gamma}_{\sigma}^{\prime }
\label{grad-pol-quad-view-mu-sig-norm-sig} \\
& =-\frac{1}{2}\boldsymbol{\gamma}_{\sigma}(\underline{\boldsymbol{\theta}}_{%
\boldsymbol{X};\sigma,\sigma}^{\mathit{N}}+\boldsymbol{\gamma}_{\sigma
}^{\prime}\boldsymbol{\theta}_{\sigma,\sigma}^{\mathit{info}}\boldsymbol{%
\gamma}_{\sigma})^{-1}\boldsymbol{\gamma}_{\sigma}^{\prime }+\boldsymbol{%
\gamma}_{\sigma}\boldsymbol{\bar{\mu}}_{\boldsymbol{X}}(\boldsymbol{\gamma}%
_{\sigma}\boldsymbol{\bar{\mu}}_{\boldsymbol{X}})^{\prime}=\boldsymbol{\eta}%
_{\sigma,\sigma}^{\mathit{info}}\text{.}  \notag
\end{align}

Let us denote the $\bar{k}_{\sigma}\times1$ updated expectation implied by
the inference input variables (\ref{view-varr}) on the second moment
conditions in (\ref{eq-mom-cond-norm})%
\begin{equation}
\boldsymbol{\eta}_{\sigma}^{\mathit{info}}\equiv\mathbb{E}^{\bar {f}_{%
\boldsymbol{X}}}\{\boldsymbol{\gamma}_{\sigma}\boldsymbol{X}\}=\boldsymbol{%
\gamma}_{\sigma}\boldsymbol{\bar{\mu}}_{\boldsymbol{X}}\text{,}
\label{implicit-view-implied}
\end{equation}
and define the following function%
\begin{equation}
\sigma^{2\mathit{info}}(\boldsymbol{\eta}_{\sigma})\equiv\boldsymbol{\eta }%
_{\sigma,\sigma}^{\mathit{info}}-\boldsymbol{\eta}_{\sigma}\boldsymbol{\eta }%
_{\sigma}^{\prime}\text{.}  \label{sig-view-func}
\end{equation}

Then, solving (\ref{grad-pol-quad-view-mu})-(\ref%
{grad-pol-quad-view-mu-sig-norm-sig}) with respect to the optimal Lagrange
multipliers (\ref{post-lagr-mult-norm}), we obtain that the optimal Lagrange
multipliers $\boldsymbol{\theta}_{\mu}^{\mathit{info}}$ are defined
implicitly in terms of $\boldsymbol{\eta}_{\sigma}^{\mathit{info}}$ (\ref%
{implicit-view-implied})%
\begin{equation}
\boldsymbol{\theta}_{\mu}^{\mathit{info}}=(\boldsymbol{\gamma}_{\mu}\bar{%
\boldsymbol{\sigma}}_{\boldsymbol{X}}^{2}\boldsymbol{\gamma}_{\mu
}^{\prime})^{-1}(\boldsymbol{\eta}_{\mu}^{\mathit{info}}-\boldsymbol{\gamma }%
_{\mu}\boldsymbol{\bar{\mu}}_{\boldsymbol{X};\sigma})\text{,}
\label{comp-quad-norm-mu}
\end{equation}
where $\boldsymbol{\bar{\mu}}_{\boldsymbol{X};\sigma}$ is the following $%
\bar{n}\times1$ vector%
\begin{align}
\boldsymbol{\bar{\mu}}_{\boldsymbol{X};\sigma} & \equiv\boldsymbol{\bar {%
\sigma}}_{\boldsymbol{X}}^{2}(\underline{\boldsymbol{\sigma}}_{\boldsymbol{X}%
}^{2})^{-1}\underline{\boldsymbol{\mu}}_{\boldsymbol{X}}  \label{mu-s} \\
& =\underline{\boldsymbol{\mu}}_{\boldsymbol{X}}+\underline{\boldsymbol{%
\sigma}}_{\boldsymbol{X}}^{2}\boldsymbol{\gamma }_{\sigma}^{\prime}(%
\boldsymbol{\gamma}_{\sigma}\underline{\boldsymbol{\sigma }}_{\boldsymbol{X}%
}^{2}\boldsymbol{\gamma}_{\sigma}^{\prime})^{-1}(\sigma^{2\mathit{info}}(%
\boldsymbol{\eta}_{\sigma}^{\mathit{info}})(\boldsymbol{\gamma}_{\sigma}%
\underline{\boldsymbol{\sigma}}_{\boldsymbol{X}}^{2}\boldsymbol{\gamma}%
_{\sigma}^{\prime})^{-1}\boldsymbol{\gamma}_{\sigma }\underline{\boldsymbol{%
\mu}}_{\boldsymbol{X}}-\boldsymbol{\gamma}_{\sigma }\underline{\boldsymbol{%
\mu}}_{\boldsymbol{X}})\text{;}  \notag
\end{align}
and similar for the optimal Lagrange multipliers $\boldsymbol{\theta}%
_{\sigma,\sigma}^{\mathit{info}}$%
\begin{equation}
\boldsymbol{\theta}_{\sigma,\sigma}^{\mathit{info}}=\frac{1}{2}((\boldsymbol{%
\gamma}_{\sigma}\underline{\boldsymbol{\sigma}}_{\boldsymbol{X}}^{2}%
\boldsymbol{\gamma}_{\sigma}^{\prime})^{-1}-(\sigma^{2\mathit{info}}(%
\boldsymbol{\eta}_{\sigma}^{\mathit{info}}))^{-1})\text{.}
\label{comp-quad-norm-sig}
\end{equation}

The above equations (\ref{sig-view-func})-(\ref{comp-quad-norm-sig}) are
implicit as long as the features $\boldsymbol{\eta}_{\sigma}^{\mathit{info}}$
(\ref{implicit-view-implied}) are \emph{not} known explicitly from the
information constraints (\ref{eq-mom-cond-norm}). For example, this
situation occurs when the rows of $\boldsymbol{\gamma}_{\sigma}$ are
linearly independent of the ones in $\boldsymbol{\gamma}_{\mu}$. Then, we
can attempt to solve numerically the equations via a fixed-point recursion,
see \cite{Colas19}.

Instead, if the rows of $\boldsymbol{\gamma}_{\sigma}$ are linearly
dependent of the ones in $\boldsymbol{\gamma}_{\mu}$, so that we can deduce $%
\boldsymbol{\eta}_{\sigma}^{\mathit{info}}$ from the known features $%
\boldsymbol{\eta}_{\mu}^{\mathit{info}}$ (\ref{grad-pol-quad-view-mu}) 
\begin{equation}
\boldsymbol{\gamma}_{\mu}\boldsymbol{\bar{\mu}}_{\boldsymbol{X}}=\boldsymbol{%
\eta}_{\mu}^{\mathit{info}}\quad\Rightarrow\quad \boldsymbol{\gamma}_{\sigma}%
\boldsymbol{\bar{\mu}}_{\boldsymbol{X}}=\boldsymbol{\eta}_{\sigma}^{\mathit{%
info}}\text{,}  \label{known-et-sig-case}
\end{equation}
the above equations (\ref{sig-view-func})-(\ref{comp-quad-norm-sig}) becomes
all explicit and hence the recursion will not be necessary. A very special
case occurs when there are constraints on expectation and covariance as in (%
\ref{view-exp-cov-norm}), as we shall see in [\ref{norm-post-pdf-quad-pol-2}%
].

\subsection{MRE gradient and Hessian with respect to expectation parameters 
\label{views-inetns-grad-hess}}

First of all, let us consider the class of exponential family distributions
as in (\ref{exp-family-set-1})%
\begin{equation}
f_{\boldsymbol{X}}^{(\boldsymbol{\eta})}(\boldsymbol{x})\equiv\underline{f}_{%
\boldsymbol{X}}(\boldsymbol{x})e^{\theta(\boldsymbol{\eta})^{\prime}\zeta(%
\boldsymbol{x})-\psi(\theta(\boldsymbol{\eta}))}\text{,}
\label{exp-set-fam-eta}
\end{equation}
for different $\bar{k}\times1$ vectors $\boldsymbol{\eta}\equiv(\eta
_{1},\ldots,\eta_{\bar{k}})^{\prime}$, where $\theta(\boldsymbol{\eta})$
denotes the link function as in (\ref{lagr-prob}) 
\begin{equation}
\theta(\boldsymbol{\eta})\equiv(\nabla_{\boldsymbol{\theta}}\psi )^{-1}(%
\boldsymbol{\eta})\text{,}  \label{max-entrpoy-link}
\end{equation}
See e.g. \cite{AmariNagaoka00} and \cite{AmariShun16} for details.

Then, the relative entropy $\mathcal{E}(f_{\boldsymbol{X}}^{(\boldsymbol{%
\eta })}\Vert\underline{f}_{\boldsymbol{X}})$ (\ref{rel-entropy}) explicitly
reads 
\begin{align}
\mathcal{E}(f_{\boldsymbol{X}}^{(\boldsymbol{\eta})}\Vert\underline{f}_{%
\boldsymbol{X}}) & =\int_{\mathcal{X}}f_{\boldsymbol{X}}^{(\boldsymbol{\eta}%
)}(\boldsymbol{x})\ln(\frac{f_{\boldsymbol{X}}^{(\boldsymbol{\eta})}(%
\boldsymbol{x})}{\underline{f}_{\boldsymbol{X}}(\boldsymbol{x})})d%
\boldsymbol{x}  \notag \\
& =\int_{\mathcal{X}}f_{\boldsymbol{X}}^{(\boldsymbol{\eta})}(\boldsymbol{x}%
)[\theta(\boldsymbol{\eta})^{\prime}\zeta(\boldsymbol{x})-\psi(\theta (%
\boldsymbol{\eta}))]d\boldsymbol{x}  \notag \\
& =\theta(\boldsymbol{\eta})^{\prime}\int_{\mathcal{X}}\zeta(\boldsymbol{x}%
)f_{\boldsymbol{X}}^{(\boldsymbol{\eta})}(\boldsymbol{x})d\boldsymbol{x}%
-\psi(\theta(\boldsymbol{\eta}))  \notag \\
& =\theta(\boldsymbol{\eta})^{\prime}\boldsymbol{\eta}-\psi(\theta (%
\boldsymbol{\eta}))\text{,}  \label{max-entrpoy}
\end{align}
where in the last row we used the fact that the exponential family
distributions $f_{\boldsymbol{X}}^{(\boldsymbol{\eta})}$ (\ref%
{exp-set-fam-eta}) satisfy by construction the information constraints on
expectations as in (\ref{view-gen-exp-norm}), or $\mathbb{E}^{f_{\boldsymbol{%
X}}^{(\boldsymbol{\eta})}}\left\{ \zeta(\boldsymbol{X})\right\} =\boldsymbol{%
\eta}$.

Then, by applying the chain rule and inverse differentiation to the link
function $\theta(\boldsymbol{\eta})$ (\ref{max-entrpoy-link}), the gradient
with respect to $\boldsymbol{\eta}$ of the relative entropy (\ref%
{max-entrpoy}) becomes the link function itself (\ref{max-entrpoy-link})%
\begin{equation}
\nabla_{\boldsymbol{\eta}}\mathcal{E}(f_{\boldsymbol{X}}^{(\boldsymbol{\eta}%
)}\Vert\underline{f}_{\boldsymbol{X}})=\theta(\boldsymbol{\eta})\text{.}
\label{max-entrpoy-grad}
\end{equation}

Moreover, by applying again the inverse differentiation to the link function 
$\theta(\boldsymbol{\eta})$ (\ref{max-entrpoy-link}), the Hessian with
respect to $\boldsymbol{\eta}$ of the relative entropy (\ref{max-entrpoy})
reads\footnote{%
The computation below fixes a minor mistake in sign for an equivalent result
in the appendix of \cite{Colas19}.} 
\begin{equation}
\nabla_{\boldsymbol{\eta},\boldsymbol{\eta}}^{2}\mathcal{E}(f_{\boldsymbol{X}%
}^{(\boldsymbol{\eta})}\Vert\underline{f}_{\boldsymbol{X}})=(\nabla _{%
\boldsymbol{\theta},\boldsymbol{\theta}}^{2}\psi(\theta(\boldsymbol{\eta }%
)))^{-1}\text{.}  \label{max-entrpoy-hess}
\end{equation}
This also means that the relative entropy $\mathcal{E}(f_{\boldsymbol{X}}^{(%
\boldsymbol{\eta})}\Vert\underline{f}_{\boldsymbol{X}})$ is a convex
function in the features $\boldsymbol{\eta}$, as follows because the
log-partition function $\psi(\boldsymbol{\theta})$ (\ref{soidfodngf}) is
also a convex function in the Lagrange multipliers $\boldsymbol{\theta}$.
See e.g. \cite{AmariNagaoka00} and \cite{AmariShun16} for details.

\subsection{MRE update under normal base and information on central moments 
\label{norm-post-pdf-quad-pol-2}}

In principle, to compute MRE solution $\bar{f}_{\boldsymbol{X}}$ (\ref%
{mre-posterior}) under information constraints on expectation and covariance 
$\mathcal{C}_{\boldsymbol{X}}$ as in (\ref{view-exp-cov-norm}), we can split
equivalently the MRE problem in two steps:\newline
i) for any given $\bar{k}_{\sigma}\times1$ vector $\boldsymbol{\eta}%
_{\sigma} $, we look at the following information constraints (\ref%
{view-exp-cov-norm-1})%
\begin{equation}
f_{\boldsymbol{X}}\in\mathcal{C}_{\boldsymbol{X}}^{(\boldsymbol{\eta}%
_{\sigma })}:\qquad\mathbb{E}^{f_{\boldsymbol{X}}}\{\left( 
\begin{matrix}
\boldsymbol{\gamma}_{\mu} & \boldsymbol{0}_{\bar{k}_{\mu}\times\bar{n}^{2}}
\\ 
\boldsymbol{\gamma}_{\sigma} & \boldsymbol{0}_{\bar{k}_{\sigma}\times\bar {n}%
^{2}} \\ 
\boldsymbol{0}_{\bar{k}_{\sigma}^{2}\times\bar{n}} & \boldsymbol{\gamma }%
_{\sigma}\otimes\boldsymbol{\gamma}_{\sigma}%
\end{matrix}
\right) \left( 
\begin{array}{c}
\boldsymbol{X} \\ 
\mathit{vec}(\boldsymbol{XX}^{\prime})%
\end{array}
\right) \}=\left( 
\begin{array}{c}
\boldsymbol{\mu}^{\mathit{info}} \\ 
\boldsymbol{\eta}_{\sigma} \\ 
\mathit{vec}(\boldsymbol{\sigma}^{2\mathit{info}}+\boldsymbol{\eta}_{\sigma }%
\boldsymbol{\eta}_{\sigma}^{\prime})%
\end{array}
\right)  \label{param-exp12-constr}
\end{equation}
and then solve the ensuing MRE problem (\ref{view-exp-cov-norm-2})%
\begin{equation}
f_{\boldsymbol{X}}^{(\boldsymbol{\eta}_{\sigma})}\equiv\limfunc{argmin}%
\limits_{f_{\boldsymbol{X}}\in\mathcal{C}_{\boldsymbol{X}}^{(\boldsymbol{%
\eta }_{\sigma})}}\mathcal{E}(f_{\boldsymbol{X}}\Vert\underline{f}_{%
\boldsymbol{X}})\text{;}  \label{first-step-an-optim}
\end{equation}
ii) we look for the optimal solution $f_{\boldsymbol{X}}^{(\boldsymbol{\eta }%
_{\sigma}^{\mathit{info}})}$ within the parametric family $\{f_{\boldsymbol{X%
}}^{(\boldsymbol{\eta}_{\sigma})}\}_{\boldsymbol{\eta }_{\sigma}}$ (\ref%
{view-exp-cov-norm-4})%
\begin{equation}
\bar{f}_{\boldsymbol{X}}=f_{\boldsymbol{X}}^{(\boldsymbol{\eta}_{\sigma }^{%
\mathit{info}})}\quad\Leftrightarrow\quad\boldsymbol{\eta}_{\sigma }^{%
\mathit{info}}\equiv\limfunc{argmin}_{\boldsymbol{\eta}_{\sigma}}\mathcal{E}%
(f_{\boldsymbol{X}}^{(\boldsymbol{\eta}_{\sigma})}\Vert \underline{f}_{%
\boldsymbol{X}})\text{.}  \label{second-step-anoptim}
\end{equation}

Now, since the information constraints $\mathcal{C}_{\boldsymbol{X}}^{(%
\boldsymbol{\eta}_{\sigma})}$ (\ref{param-exp12-constr}) are statements on
the first two non-central moments of the target variables (\ref%
{eq-mom-cond-norm}), under normality of the base $\underline{f}_{\boldsymbol{%
X}}$ (\ref{norm-exp-form}) the ensuing updated distribution $f_{\boldsymbol{X%
}}^{(\boldsymbol{\eta}_{\sigma})}$ (\ref{first-step-an-optim}) must be in
the same exponential family class of the base (\ref{norm-exp-form}), and
hence normal in turn (\ref{norm-exp-form}).

In particular, under the information constraints (\ref{param-exp12-constr}),
the updated covariance function of the inference input variables $\sigma^{2%
\mathit{info}}(\boldsymbol{\eta}_{\sigma})$ (\ref{sig-view-func}) is
constant in each fixed $\boldsymbol{\eta}_{\sigma}$%
\begin{equation}
\sigma^{2\mathit{info}}(\boldsymbol{\eta}_{\sigma})=(\boldsymbol{\sigma }^{2%
\mathit{info}}+\boldsymbol{\eta}_{\sigma}\boldsymbol{\eta}_{\sigma
}^{\prime})-\boldsymbol{\eta}_{\sigma}\boldsymbol{\eta}_{\sigma}^{\prime }=%
\boldsymbol{\sigma}^{2\mathit{info}}\text{,}  \label{second-step-anoptim-1}
\end{equation}
and hence the optimal Lagrange multipliers $\boldsymbol{\theta}_{\sigma
,\sigma}^{\mathit{info}}$ (\ref{comp-quad-norm-sig}), as well as the updated
covariance $\boldsymbol{\bar{\sigma}}_{\boldsymbol{X}}^{2}$ (\ref%
{pos-exp-nomr-2-sig}), must be explicit in turn.

Indeed, the optimal Lagrange multipliers $\boldsymbol{\theta}_{\sigma,\sigma
}^{\mathit{info}}$ (\ref{comp-quad-norm-sig}) becomes 
\begin{equation}
\boldsymbol{\theta}_{\sigma,\sigma}^{\mathit{info}}=\frac{1}{2}((\boldsymbol{%
\gamma}_{\sigma}\underline{\boldsymbol{\sigma}}_{\boldsymbol{X}}^{2}%
\boldsymbol{\gamma}_{\sigma}^{\prime})^{-1}-(\boldsymbol{\sigma }^{2\mathit{%
info}})^{-1})\text{;}  \label{comp-quad-norm-sig-2}
\end{equation}
and using the the binomial inverse theorem \cite{magnus79}, it is immediate
that the updated covariance $\boldsymbol{\bar{\sigma}}_{\boldsymbol{X}}^{2}$
(\ref{pos-exp-nomr-2-sig})\ becomes as in (\ref{comp-quad-norm-sig-1}).

Instead, if we define the following matrix $(\bar{k}_{\mu}+\bar{k}_{\sigma
})\times\bar{n}$ matrix%
\begin{equation}
\tilde{\boldsymbol{\gamma}}_{\mu}\equiv\left( 
\begin{matrix}
\boldsymbol{\gamma}_{\mu} \\ 
\boldsymbol{\gamma}_{\sigma}%
\end{matrix}
\right) \text{,}  \label{gamm-mu-sig}
\end{equation}
the other vector of Lagrange multipliers as in (\ref{comp-quad-norm-mu}) is
trivially explicit in each fixed $\boldsymbol{\eta}_{\sigma}$ 
\begin{equation}
\left( 
\begin{matrix}
\theta_{\mu}(\boldsymbol{\eta}_{\sigma}) \\ 
\theta_{\sigma}(\boldsymbol{\eta}_{\sigma})%
\end{matrix}
\right) \equiv(\tilde{\boldsymbol{\gamma}}_{\mu}\boldsymbol{\bar{\sigma}}_{%
\boldsymbol{X}}^{2}\tilde{\boldsymbol{\gamma}}_{\mu}^{\prime})^{-1}\left( 
\begin{array}{c}
\boldsymbol{\mu}^{\mathit{info}}-\boldsymbol{\gamma}_{\mu}\bar{\boldsymbol{%
\mu }}_{\boldsymbol{X};\sigma} \\ 
\boldsymbol{\eta}_{\sigma}-\boldsymbol{\gamma}_{\sigma}\boldsymbol{\bar{\mu}}%
_{\boldsymbol{X};\sigma}%
\end{array}
\right) \text{,}  \label{thet-mu-sig}
\end{equation}
as well as the ensuing updated expectation as in (\ref{mu-yet})%
\begin{equation}
\mu(\boldsymbol{\eta}_{\sigma})\equiv\boldsymbol{\bar{\sigma}}_{\boldsymbol{X%
}}^{2}(\underline{\boldsymbol{\theta}}_{\boldsymbol{X};\mu}^{\mathit{N}}+%
\boldsymbol{\gamma}_{\mu}^{\prime}\theta_{\mu}(\boldsymbol{\eta}_{\sigma })+%
\boldsymbol{\gamma}_{\sigma}^{\prime}\theta_{\sigma}(\boldsymbol{\eta }%
_{\sigma}))\text{.}  \label{mu-yet-mu-sig}
\end{equation}

Now, since gradient of the relative entropy objective in (\ref%
{second-step-anoptim}) here reads [\ref{views-inetns-grad-hess}]%
\begin{equation}
\nabla_{\boldsymbol{\eta}_{\sigma}}\mathcal{E}(f_{\boldsymbol{X}}^{(%
\boldsymbol{\eta}_{\sigma})}\Vert\underline{f}_{\boldsymbol{X}%
})=\theta_{\sigma}(\boldsymbol{\eta}_{\sigma})\text{,}
\label{grad-entrp-eta-sig}
\end{equation}
then the optimal $\boldsymbol{\eta}_{\sigma}^{\mathit{info}}$ in (\ref%
{second-step-anoptim}) must solve the following first order conditions%
\begin{equation}
\theta_{\sigma}(\boldsymbol{\eta}_{\sigma}^{\mathit{info}})\equiv 
\boldsymbol{0}_{\bar{k}_{\sigma}\times1}\text{.}
\label{grad-entrp-eta-sig-1}
\end{equation}
Note how the Lagrange multipliers $\theta_{\sigma}(\boldsymbol{\eta}_{\sigma
})$ (\ref{thet-mu-sig}) are increasing linear functions in $\boldsymbol{\eta 
}_{\sigma}$, and hence from (\ref{grad-entrp-eta-sig}), the relative entropy
objective $\mathcal{E}(f_{\boldsymbol{X}}^{(\boldsymbol{\eta}_{\sigma})}\Vert%
\underline{f}_{\boldsymbol{X}})$ in (\ref{second-step-anoptim}) must be a
convex quadratic function in $\boldsymbol{\eta}_{\sigma}$.

To solve the above, let us first arrange in blocks the symmetric $(\bar {k}%
_{\mu}+\bar{k}_{\sigma})\times(\bar{k}_{\mu}+\bar{k}_{\sigma})$ matrix $(%
\tilde{\boldsymbol{\gamma}}_{\mu}\boldsymbol{\bar{\sigma}}_{\boldsymbol{X}%
}^{2}\tilde{\boldsymbol{\gamma}}_{\mu}^{\prime})^{-1}$ in (\ref{thet-mu-sig}%
) as follows%
\begin{equation}
\left( 
\begin{matrix}
\boldsymbol{\omega}_{\mu,\mu} & \boldsymbol{\omega}_{\sigma,\mu}^{\prime} \\ 
\boldsymbol{\omega}_{\sigma,\mu} & \boldsymbol{\omega}_{\sigma,\sigma}%
\end{matrix}
\right) \equiv\left( 
\begin{matrix}
\boldsymbol{\gamma}_{\mu}\boldsymbol{\bar{\sigma}}_{\boldsymbol{X}}^{2}%
\boldsymbol{\gamma}_{\mu}^{\prime} & \boldsymbol{\gamma}_{\mu}\bar{%
\boldsymbol{\sigma}}_{\boldsymbol{X}}^{2}\boldsymbol{\gamma}_{\sigma
}^{\prime} \\ 
\boldsymbol{\gamma}_{\sigma}\boldsymbol{\bar{\sigma}}_{\boldsymbol{X}}^{2}%
\boldsymbol{\gamma}_{\mu}^{\prime} & \boldsymbol{\gamma}_{\sigma}\bar{%
\boldsymbol{\sigma}}_{\boldsymbol{X}}^{2}\boldsymbol{\gamma}_{\sigma
}^{\prime}%
\end{matrix}
\right) ^{-1}=(\tilde{\boldsymbol{\gamma}}_{\mu}\boldsymbol{\bar{\sigma}}_{%
\boldsymbol{X}}^{2}\tilde{\boldsymbol{\gamma}}_{\mu}^{\prime})^{-1}\text{.}
\label{block-inv-sig-evid}
\end{equation}

Then, using (\ref{thet-mu-sig})-(\ref{block-inv-sig-evid}), the first order
conditions (\ref{grad-entrp-eta-sig-1}) becomes 
\begin{equation}
\boldsymbol{\omega}_{\sigma,\mu}(\boldsymbol{\mu}^{\mathit{info}}-%
\boldsymbol{\gamma}_{\mu}\boldsymbol{\bar{\mu}}_{\boldsymbol{X};\sigma })+%
\boldsymbol{\omega}_{\sigma,\sigma}(\boldsymbol{\eta}_{\sigma }^{\mathit{info%
}}-\boldsymbol{\gamma}_{\sigma}\boldsymbol{\bar{\mu}}_{\boldsymbol{X}%
;\sigma})=\boldsymbol{0}_{\bar{k}_{\sigma}\times1}\text{,}
\label{grad-entrp-eta-sig-2}
\end{equation}
which implies%
\begin{align}
\boldsymbol{\eta}_{\sigma}^{\mathit{info}} & =\boldsymbol{\gamma}_{\sigma }%
\boldsymbol{\bar{\mu}}_{\boldsymbol{X};\sigma}-\boldsymbol{\omega}%
_{\sigma,\sigma}^{-1}\boldsymbol{\omega}_{\sigma,\mu}(\boldsymbol{\mu }^{%
\mathit{info}}-\boldsymbol{\gamma}_{\mu}\boldsymbol{\bar{\mu}}_{\boldsymbol{X%
};\sigma})  \label{opt-eta-sig} \\
& =\boldsymbol{\gamma}_{\sigma}\boldsymbol{\bar{\mu}}_{\boldsymbol{X};\sigma
}+\boldsymbol{\gamma}_{\sigma}\boldsymbol{\bar{\sigma}}_{\boldsymbol{X}}^{2}%
\boldsymbol{\gamma}_{\mu}^{\prime}(\boldsymbol{\gamma}_{\mu}\boldsymbol{\bar{%
\sigma}}_{\boldsymbol{X}}^{2}\boldsymbol{\gamma}_{\mu }^{\prime})^{-1}(%
\boldsymbol{\mu}^{\mathit{info}}-\boldsymbol{\gamma}_{\mu }\boldsymbol{\bar{%
\mu}}_{\boldsymbol{X};\sigma})  \notag \\
& =\boldsymbol{\sigma}^{2\mathit{info}}(\boldsymbol{\gamma}_{\sigma }%
\underline{\boldsymbol{\sigma}}_{\boldsymbol{X}}^{2}\boldsymbol{\gamma }%
_{\sigma}^{\prime})^{-1}\boldsymbol{\gamma}_{\sigma}\underline{\boldsymbol{%
\mu}}_{\boldsymbol{X}}+\boldsymbol{\gamma}_{\sigma}\bar{\boldsymbol{\gamma}}%
_{\mu}^{\dag}(\boldsymbol{\mu}^{\mathit{info}}-\boldsymbol{\gamma}_{\mu}%
\boldsymbol{\bar{\mu}}_{\boldsymbol{X};\sigma })\text{,}  \notag
\end{align}
where the second row follows from the block matrix inversion applied to (\ref%
{block-inv-sig-evid}), i.e.%
\begin{equation}
\boldsymbol{\omega}_{\sigma,\mu}=-\boldsymbol{\omega}_{\sigma,\sigma }(%
\boldsymbol{\gamma}_{\sigma}\boldsymbol{\bar{\sigma}}_{\boldsymbol{X}}^{2}%
\boldsymbol{\gamma}_{\mu}^{\prime})(\boldsymbol{\gamma}_{\mu}\bar{%
\boldsymbol{\sigma}}_{\boldsymbol{X}}^{2}\boldsymbol{\gamma}_{\mu
}^{\prime})^{-1}\text{,}  \label{opt-eta-sig-1}
\end{equation}
see \cite{magnus79}; and the third row follows by definition of the
pseudo-inverse $\bar{\boldsymbol{\gamma}}_{\mu}^{\dag}$ (\ref%
{mu-pseudo-inv-pos}) and 
\begin{equation}
\boldsymbol{\gamma}_{\sigma}\boldsymbol{\bar{\mu}}_{\boldsymbol{X};\sigma }=%
\boldsymbol{\sigma}^{2\mathit{info}}(\boldsymbol{\gamma}_{\sigma }\underline{%
\boldsymbol{\sigma}}_{\boldsymbol{X}}^{2}\boldsymbol{\gamma }%
_{\sigma}^{\prime})^{-1}\boldsymbol{\gamma}_{\sigma}\underline{\boldsymbol{%
\mu}}_{\boldsymbol{X}}\text{,}  \label{opt-eta-sig-2}
\end{equation}
as follows from $\boldsymbol{\bar{\mu}}_{\boldsymbol{X};\sigma}$ (\ref{mu-s}%
).

This implies, that the optimal Lagrange multipliers $\boldsymbol{\theta}%
_{\mu }^{\mathit{info}}\equiv\theta_{\mu}(\boldsymbol{\eta}_{\sigma}^{%
\mathit{info}})$ in (\ref{thet-mu-sig}) becomes 
\begin{align}
\theta_{\mu}(\boldsymbol{\eta}_{\sigma}^{\mathit{info}}) & =\boldsymbol{%
\omega}_{\mu,\mu}(\boldsymbol{\mu}^{\mathit{info}}-\boldsymbol{\gamma}_{\mu}%
\boldsymbol{\bar{\mu}}_{\boldsymbol{X};\sigma })+\boldsymbol{\omega}%
_{\sigma,\mu}^{\prime}(\boldsymbol{\eta}_{\sigma }^{\mathit{info}}-%
\boldsymbol{\gamma}_{\sigma}\boldsymbol{\bar{\mu}}_{\boldsymbol{X};\sigma}) 
\notag \\
& =\boldsymbol{\omega}_{\mu,\mu}(\boldsymbol{\mu}^{\mathit{info}}-%
\boldsymbol{\gamma}_{\mu}\boldsymbol{\bar{\mu}}_{\boldsymbol{X};\sigma })+%
\boldsymbol{\omega}_{\sigma,\mu}^{\prime}(\boldsymbol{\gamma}_{\sigma }%
\boldsymbol{\bar{\sigma}}_{\boldsymbol{X}}^{2}\boldsymbol{\gamma}_{\mu
}^{\prime})(\boldsymbol{\gamma}_{\mu}\boldsymbol{\bar{\sigma}}_{\boldsymbol{X%
}}^{2}\boldsymbol{\gamma}_{\mu}^{\prime})^{-1}(\boldsymbol{\mu}^{\mathit{info%
}}-\boldsymbol{\gamma}_{\mu}\boldsymbol{\bar{\mu}}_{\boldsymbol{X};\sigma })
\notag \\
& =(\boldsymbol{\omega}_{\mu,\mu}+\boldsymbol{\omega}_{\sigma,\mu}^{\prime }(%
\boldsymbol{\gamma}_{\sigma}\boldsymbol{\bar{\sigma}}_{\boldsymbol{X}}^{2}%
\boldsymbol{\gamma}_{\mu}^{\prime})(\boldsymbol{\gamma}_{\mu }\boldsymbol{%
\bar{\sigma}}_{\boldsymbol{X}}^{2}\boldsymbol{\gamma}_{\mu }^{\prime})^{-1})(%
\boldsymbol{\mu}^{\mathit{info}}-\boldsymbol{\gamma}_{\mu }\boldsymbol{\bar{%
\mu}}_{\boldsymbol{X};\sigma})  \notag \\
& =(\boldsymbol{\gamma}_{\mu}\boldsymbol{\bar{\sigma}}_{\boldsymbol{X}}^{2}%
\boldsymbol{\gamma}_{\mu}^{\prime})^{-1}(\boldsymbol{\mu}^{\mathit{info}}-%
\boldsymbol{\gamma}_{\mu}\boldsymbol{\bar{\mu}}_{\boldsymbol{X};\sigma })%
\text{,}  \label{thet-mu-sig-1}
\end{align}
where the last row follows from the block matrix inversion applied to (\ref%
{block-inv-sig-evid}), i.e. the binomial inverse theorem%
\begin{equation}
\boldsymbol{\omega}_{\mu,\mu}=(\boldsymbol{\gamma}_{\mu}\bar {\boldsymbol{%
\sigma}}_{\boldsymbol{X}}^{2}\boldsymbol{\gamma}_{\mu}^{\prime })^{-1}-%
\boldsymbol{\omega}_{\sigma,\mu}^{\prime}(\boldsymbol{\gamma}_{\sigma }%
\boldsymbol{\bar{\sigma}}_{\boldsymbol{X}}^{2}\boldsymbol{\gamma}_{\mu
}^{\prime})(\boldsymbol{\gamma}_{\mu}\boldsymbol{\bar{\sigma}}_{\boldsymbol{X%
}}^{2}\boldsymbol{\gamma}_{\mu}^{\prime})^{-1}\text{,}  \label{thet-mu-sig-2}
\end{equation}
see \cite{magnus79}.

Therefore the optimal vector of Lagrange multipliers (\ref{thet-mu-sig})
becomes%
\begin{equation}
\left( 
\begin{matrix}
\boldsymbol{\theta}_{\mu}^{\mathit{info}} \\ 
\boldsymbol{\theta}_{\sigma}^{\mathit{info}}%
\end{matrix}
\right) =\left( 
\begin{matrix}
\theta_{\mu}(\boldsymbol{\eta}_{\sigma}^{\mathit{info}}) \\ 
\theta_{\sigma}(\boldsymbol{\eta}_{\sigma}^{\mathit{info}})%
\end{matrix}
\right) =\left( 
\begin{matrix}
(\boldsymbol{\gamma}_{\mu}\boldsymbol{\bar{\sigma}}_{\boldsymbol{X}}^{2}%
\boldsymbol{\gamma}_{\mu}^{\prime})^{-1}(\boldsymbol{\mu}^{\mathit{info}}-%
\boldsymbol{\gamma}_{\mu}\boldsymbol{\bar{\mu}}_{\boldsymbol{X};\sigma}) \\ 
\boldsymbol{0}_{\bar{k}_{\sigma}\times1}%
\end{matrix}
\right) \text{,}  \label{comp-quad-norm-mu-2}
\end{equation}
from which follows that the updated expectation $\boldsymbol{\bar{\mu}}_{%
\boldsymbol{X}}\equiv\mu(\boldsymbol{\eta}_{\sigma}^{\mathit{info}})$ (\ref%
{mu-yet-mu-sig}) becomes as in (\ref{comp-quad-norm-mu-1}).

\subsection{MRE update under normal base and uncorrelated inference
variables \label{norm-post-pdf-quad-pol-3}}

According to the results in [\ref{norm-post-pdf-quad-pol-2}], if the
inference input variables are uncorrelated under the base (\ref{uncorr-view}%
), then we must have%
\begin{align}
\boldsymbol{\gamma}_{\mu}\boldsymbol{\bar{\mu}}_{\boldsymbol{X};\sigma} & =%
\underline{\boldsymbol{\mu}}_{\boldsymbol{X}}+\boldsymbol{\gamma}_{\sigma
}^{\dag}(\boldsymbol{\sigma}^{2\mathit{info}}(\boldsymbol{\gamma}_{\sigma }%
\underline{\boldsymbol{\sigma}}_{\boldsymbol{X}}^{2}\boldsymbol{\gamma }%
_{\sigma}^{\prime})^{-1}\boldsymbol{\gamma}_{\sigma}\underline{\boldsymbol{%
\mu}}_{\boldsymbol{X}}-\boldsymbol{\gamma}_{\sigma }\underline{\boldsymbol{%
\mu}}_{\boldsymbol{X}})  \notag \\
& =\boldsymbol{\gamma}_{\mu}\underline{\boldsymbol{\mu}}_{\boldsymbol{X}}%
\text{,}  \label{comp-quad-norm-mu-3}
\end{align}
as follows because 
\begin{equation}
\boldsymbol{\gamma}_{\mu}\boldsymbol{\gamma}_{\sigma}^{\dag}=\boldsymbol{%
\gamma}_{\mu}\underline{\boldsymbol{\sigma}}_{\boldsymbol{X}}^{2}\boldsymbol{%
\gamma}_{\sigma}^{\prime}(\boldsymbol{\gamma}_{\sigma }\underline{%
\boldsymbol{\sigma}}_{\boldsymbol{X}}^{2}\boldsymbol{\gamma }%
_{\sigma}^{\prime})^{-1}=\boldsymbol{0}_{\bar{k}_{\mu}\times\bar{k}_{\sigma}}%
\text{.}  \label{comp-quad-norm-mu-4}
\end{equation}
This implies the optimal Lagrange multipliers $\boldsymbol{\theta}_{\mu }^{%
\mathit{info}}$ (\ref{comp-quad-norm-mu-2}) simplifies to%
\begin{equation}
\boldsymbol{\theta}_{\mu}^{\mathit{info}}=(\boldsymbol{\gamma}_{\mu }%
\underline{\boldsymbol{\sigma}}_{\boldsymbol{X}}^{2}\boldsymbol{\gamma}_{\mu
}^{\prime})^{-1}(\boldsymbol{\mu}^{\mathit{info}}-\boldsymbol{\gamma}_{\mu }%
\underline{\boldsymbol{\mu}}_{\boldsymbol{X}})\text{,}
\label{comp-quad-norm-mu-5}
\end{equation}
and the updated expectation $\boldsymbol{\bar{\mu}}_{\boldsymbol{X}}$ (\ref%
{comp-quad-norm-mu-1}) becomes as in (\ref{pseudo-inv-propp-4}).

Finally, the pseudo inverse (\ref{mu-pseudo-inv-pos}) becomes%
\begin{equation}
\bar{\boldsymbol{\gamma}}_{\mu}^{\dag}=\boldsymbol{\gamma}_{\mu}^{\dag}\equiv%
\underline{\boldsymbol{\sigma}}_{\boldsymbol{X}}^{2}\boldsymbol{\gamma }%
_{\mu}^{\prime}(\boldsymbol{\gamma}_{\sigma}\underline{\boldsymbol{\sigma}}_{%
\boldsymbol{X}}^{2}\boldsymbol{\gamma}_{\sigma}^{\prime})^{-1}\text{,}
\label{comp-quad-norm-mu-6}
\end{equation}
as follows by replacing the explicit expression of the updated covariance $%
\boldsymbol{\bar{\sigma}}_{\boldsymbol{X}}^{2}$ (\ref{pos-exp-nomr-2-sig})
and using the orthogonality condition (\ref{comp-quad-norm-mu-4}).

\subsection{HMC sampling for exponential family distributions\label%
{hmc-smapling-sec}}

The Hamiltonian Monte Carlo (HMC) sampling approach \cite{Chao15}, \cite%
{Neal11}, is a Markov chain Monte Carlo (MCMC) method and as such, it is
unaffected by scaling, i.e. it allows to sample from an arbitrary
distribution $f_{\boldsymbol{X}}$ of the form 
\begin{equation}
f_{\boldsymbol{X}}(\boldsymbol{x})\propto g(\boldsymbol{x})\text{,}
\label{scale-inv}
\end{equation}
with the only knowledge of $g(\boldsymbol{x})$. This is particularly useful
to sample from an exponential family distribution%
\begin{equation}
f_{\boldsymbol{X}}\propto\underline{g}_{\boldsymbol{X}}(\boldsymbol{x})e^{%
\boldsymbol{\theta}^{\prime}\zeta(\boldsymbol{x})}\text{,}
\label{scale-inv-1}
\end{equation}
for a given $\boldsymbol{\theta}$, including both base distribution (\ref%
{generic-prior}) (case $\boldsymbol{\theta}=\boldsymbol{0}$) and updated
counterpart (\ref{generic-post-1}) (case $\boldsymbol{\theta}=\boldsymbol{%
\theta}^{\mathit{info}}$).

More precisely, the HMC algorithm needs two inputs:\newline
i) the log-pdf modulo constant terms, which here reads%
\begin{equation}
u(\boldsymbol{x})\equiv\boldsymbol{\theta}^{\prime}\zeta(\boldsymbol{x})+\ln%
\underline{f}_{\boldsymbol{X}}(\boldsymbol{x})\text{;}  \label{pt-energy}
\end{equation}
ii) (optionally) the respective gradient, which here reads%
\begin{equation}
\nabla_{\boldsymbol{x}}u(\boldsymbol{x})=\frac{1}{\underline{f}_{\boldsymbol{%
X}}(\boldsymbol{x})}\nabla_{\boldsymbol{x}}\underline{f}_{\boldsymbol{X}}(%
\boldsymbol{x})+J_{\zeta}(\boldsymbol{x})^{\prime }\boldsymbol{\theta}\text{,%
}  \label{grad-energy}
\end{equation}
and where $J_{\zeta}(\boldsymbol{x})$ denotes the $\bar{k}\times\bar{n}$
Jacobian matrix of the information function $\zeta(\boldsymbol{x})$.

Indeed, the generic $n$-th partial derivative of the log-pdf (\ref{pt-energy}%
) reads%
\begin{align}
\lbrack\nabla_{\boldsymbol{x}}u(\boldsymbol{x})]_{n} & \equiv\frac{\partial 
}{\partial x_{n}}u(\boldsymbol{x})=\frac{\partial}{\partial x_{n}}[\tsum
\nolimits_{k=1}^{\bar{k}}\theta_{k}\zeta_{k}(\boldsymbol{x})+\frac{\partial}{%
\partial x_{n}}\ln\underline{f}_{\boldsymbol{X}}(\boldsymbol{x})]  \notag \\
& =\tsum \nolimits_{k=1}^{\bar{k}}\theta_{k}\frac{\partial}{\partial x_{n}}%
\zeta_{k}(\boldsymbol{x})+\frac {1}{\underline{f}_{\boldsymbol{X}}(%
\boldsymbol{x})}\frac{\partial}{\partial x_{n}}\underline{f}_{\boldsymbol{X}%
}(\boldsymbol{x})  \notag \\
& =\tsum \nolimits_{k=1}^{\bar{k}}\theta_{k}[J_{\zeta}(\boldsymbol{x}%
)]_{k,n}+\frac{1}{\underline{f}_{\boldsymbol{X}}(\boldsymbol{x})}[\nabla_{%
\boldsymbol{x}}\underline{f}_{\boldsymbol{X}}(\boldsymbol{x})]_{n}  \notag \\
& =[J_{\zeta}(\boldsymbol{x})^{\prime}\boldsymbol{\theta}]_{n}+\frac {1}{%
\underline{f}_{\boldsymbol{X}}(\boldsymbol{x})}[\nabla_{\boldsymbol{x}}%
\underline{f}_{\boldsymbol{X}}(\boldsymbol{x})]_{n}\text{,}
\label{grad-energy-1}
\end{align}
where in the third row we used the definition of Jacobian matrix%
\begin{equation}
\lbrack J_{\zeta}(\boldsymbol{x})]_{k,n}\equiv\frac{\partial}{\partial x_{n}}%
\zeta_{k}(\boldsymbol{x})\text{.}  \label{grad-energy-2}
\end{equation}
Hence comparing both sides of the above identity we obtain the desired
result (\ref{grad-energy}).

\subsection{Exponential invariance of the updated distribution\label%
{composition-appx-1}}

Suppose that our base distribution (\ref{prior-distr}) is within an
exponential family class as in (\ref{exp-fam-base})%
\begin{equation}
\underline{f}_{\boldsymbol{X}}\sim\mathit{Exp}(\underline{\boldsymbol{\theta}%
},\zeta,h,\mathcal{X})\text{,}  \label{pos-wrt-base-case-0}
\end{equation}
for some base vector $\underline{\boldsymbol{\theta}}\equiv(\underline{%
\theta }_{1},\ldots,\underline{\theta}_{\bar{k}})^{\prime}\in\Theta$ of
canonical coordinates and arbitrary reference measure $h(\boldsymbol{x})>0$,
which without loss of generality we can assume to be normalized $\int h(%
\boldsymbol{x})d\boldsymbol{x}=1$. Note that this case includes the original
base in (\ref{generic-prior}) as well as the new one in (\ref{new-estim-base}%
).

Generalizing results in [\ref{gen-post-pdf-appendix}], under information
conditions on generalized expectations $\mathcal{C}_{\boldsymbol{X}}$ (\ref%
{view-gen-exp-norm}), the MRE updated distribution $\bar{f}_{\boldsymbol{X}%
}\sim\mathit{Exp}(\boldsymbol{\theta}^{\mathit{info}},\zeta,\underline{f}_{%
\boldsymbol{X}},\mathcal{X})$ (\ref{exp-family-set}) can be represented as
an exponential family distribution under the reference measure $h$, as long
as $\underline{\boldsymbol{\theta}}+\boldsymbol{\theta }^{\mathit{info}%
}\in\Theta$%
\begin{equation}
\bar{f}_{\boldsymbol{X}}\sim\mathit{Exp}(\underline{\boldsymbol{\theta}}+%
\boldsymbol{\theta}^{\mathit{info}},\zeta,h,\mathcal{X})\text{,}
\label{pos-wrt-base-case}
\end{equation}
where the original log-partition function $\psi_{\underline{f}_{\boldsymbol{X%
}},\zeta}$ (\ref{soidfodngf}) can be written in terms of the log-partition
function $\psi_{h,\zeta}$ under the reference measure $h$ (\ref{gen-log-part}%
) as follows%
\begin{equation}
\psi_{\underline{f}_{\boldsymbol{X}},\zeta}(\boldsymbol{\theta})=\psi
_{h,\zeta}(\underline{\boldsymbol{\theta}}+\boldsymbol{\theta}%
)-\psi_{h,\zeta }(\underline{\boldsymbol{\theta}})\text{.}
\label{sum-gen-logs-3}
\end{equation}
Moreover, $\bar{f}_{\boldsymbol{X}}$ (\ref{pos-wrt-base-case}) must be also
the MRE updated distribution (\ref{mre-posterior}) under the same
information conditions $\mathcal{C}_{\boldsymbol{X}}$ (\ref%
{view-gen-exp-norm}), and reference measure $h$ as base distribution (\ref%
{prior-distr}).

To this purpose, we just need to verify that the vector $\underline{%
\boldsymbol{\theta}}+\boldsymbol{\theta}^{\mathit{info}}$ is the solution of
the dual Lagrangian problem (\ref{lagr-prob}) 
\begin{equation}
\underline{\boldsymbol{\theta}}+\boldsymbol{\theta}^{\mathit{info}}=\limfunc{%
argmin}\limits_{\boldsymbol{\vartheta}}\psi_{h,\zeta }(\boldsymbol{\vartheta}%
)-\boldsymbol{\vartheta}^{\prime}\boldsymbol{\eta }^{\mathit{info}}\text{.}
\label{pos-wrt-base-case-1}
\end{equation}

Indeed, the original dual Lagrangian problem (\ref{lagr-prob}) is equivalent
to%
\begin{align}
\boldsymbol{\theta}^{\mathit{info}} & \equiv\limfunc{argmin}\limits_{%
\boldsymbol{\theta}}\psi_{\underline{f}_{\boldsymbol{X}},\zeta }(\boldsymbol{%
\theta})-\boldsymbol{\theta}^{\prime}\boldsymbol{\eta }^{\mathit{info}} 
\notag \\
& =\limfunc{argmin}\limits_{\boldsymbol{\theta}}\psi_{h,\zeta }(\underline{%
\boldsymbol{\theta}}+\boldsymbol{\theta})-\psi_{h,\zeta }(\underline{%
\boldsymbol{\theta}})-\boldsymbol{\theta}^{\prime}\boldsymbol{\eta}^{\mathit{%
info}}  \notag \\
& =\limfunc{argmin}\limits_{\boldsymbol{\theta}}\psi_{h,\zeta }(\underline{%
\boldsymbol{\theta}}+\boldsymbol{\theta})-(\underline{\boldsymbol{\theta}}+%
\boldsymbol{\theta})^{\prime}\boldsymbol{\eta}^{\mathit{info}}\text{,}
\label{pos-wrt-base-case-2}
\end{align}
where the second row follows from (\ref{sum-gen-logs-3}); and the last row
follows because the constant terms $\psi_{h,\zeta}(\underline{\boldsymbol{%
\theta}})$ and $\underline{\boldsymbol{\theta}}^{\prime}\boldsymbol{\eta}^{%
\mathit{info}}$ do not alter the optimization problem.

Hence, changing the coordinates $\boldsymbol{\theta }$ in (\ref%
{pos-wrt-base-case-1}) by shifting%
\begin{equation}
\boldsymbol{\vartheta }\equiv \underline{\boldsymbol{\theta }}+\boldsymbol{%
\theta }\text{,}  \label{pos-wrt-base-case-3}
\end{equation}%
we obtain the desired result (\ref{pos-wrt-base-case}).

\end{document}